\documentclass[runningheads]{llncs}

\usepackage[mobile]{eccv}

\usepackage{eccvabbrv}

\usepackage{graphicx}
\usepackage{booktabs}

\usepackage{amsmath}
\usepackage{amssymb}
\usepackage{xcolor}
\usepackage{color}
\usepackage{bbding}
\usepackage{array,multirow,textcomp}
\usepackage{rotating}
\usepackage[skip=0.5\baselineskip]{caption}
\usepackage{makecell}
\usepackage{booktabs}
\usepackage{adjustbox}
\usepackage{array}
\usepackage{balance}
\usepackage{wrapfig}
\usepackage{enumitem}
\usepackage{tikz}
\usepackage{tabularx}
\usepackage{colortbl}
\usepackage{placeins}
\usepackage{float}
\usepackage{pgf}
\usepackage{subcaption}

\newcolumntype{Y}{>{\centering\arraybackslash}X}
\newcolumntype{C}[1]{>{\centering\let\newline\\\arraybackslash\hspace{0pt}}m{#1}}
\newcolumntype{R}[2]{%
    >{\adjustbox{angle=#1,lap=\width-(#2)}\bgroup}%
    l%
    <{\egroup}%
}
\newcommand{\tabincell}[2]{\begin{tabular}{@{}#1@{}}#2\end{tabular}}
\newcommand*\rots{\multicolumn{1}{R{60}{1em}}}

\usepackage[dvipsnames]{xcolor}

\usepackage{pifont}
\newcommand{\cmark}{\ding{51}}%
\newcommand{\xmark}{\ding{55}}%

\usepackage{tikz}
\newcommand*\circled[1]{\tikz[baseline=(char.base)]{
\node[shape=circle,fill=black,inner sep=0.5pt] (char) {\textcolor{white}{\scriptsize \textbf{#1}}};}}

\makeatletter
\DeclareRobustCommand\onedot{\futurelet\@let@token\@onedot}
\def\@onedot{\ifx\@let@token.\else.\null\fi\xspace}

\def\eg{\emph{e.g}\onedot} 
\def\ie{\emph{i.e}\onedot}

\def\etal{\emph{et al}\onedot}
\makeatother

\usepackage[accsupp]{axessibility}  

\usepackage{hyperref}

\usepackage{orcidlink}

\makeatletter
\renewcommand*{\@fnsymbol}[1]{\ensuremath{\ifcase#1\or *\or \dagger\or \ddagger\or
    \mathsection\or \mathparagraph\or \|\or **\or \dagger\dagger
    \or \ddagger\ddagger \else\@ctrerr\fi}}
\makeatother

\begin{document}

\title{Occlusion-Aware Seamless Segmentation} 

\titlerunning{OASS}

\author{Yihong Cao\inst{1,}\thanks{Equal contribution}\orcidlink{0000-0003-1751-5505} 
\and Jiaming Zhang\inst{2,*}\orcidlink{0000-0003-3471-328X} 
\and Hao Shi\inst{3}\orcidlink{0000-0003-0184-2245} 
\and Kunyu Peng\inst{2}\orcidlink{0000-0002-5419-9292} 
\and\\Yuhongxuan Zhang\inst{1}\orcidlink{0009-0001-6671-3754} 
\and Hui Zhang\inst{1,}\thanks{Correspondence: zhanghuihby@126.com, kailun.yang@hnu.edu.cn}\orcidlink{0000-0002-1803-3148}  
\and Rainer Stiefelhagen\inst{2}\orcidlink{0000-0001-8046-4945}
\and\\Kailun Yang\inst{1,\dagger}\orcidlink{0000-0002-1090-667X}
}

\authorrunning{Y.~Cao, J.~Zhang et al.}

\institute{$^1$Hunan University, $^2$Karlsruhe Institute of Technology, $^3$Zhejiang University}

\maketitle

\begin{abstract}
 Panoramic images can broaden the Field of View (FoV), occlusion-aware prediction can deepen the understanding of the scene, and domain adaptation can transfer across viewing domains. In this work, we introduce a novel task, \textbf{Occlusion-Aware Seamless Segmentation (OASS)}, which simultaneously tackles all these three challenges. For benchmarking OASS, we establish a new human-annotated dataset for Blending Panoramic Amodal Seamless Segmentation, \ie, \textbf{BlendPASS}. Besides, we propose the first solution \textbf{UnmaskFormer}, aiming at unmasking the narrow FoV, occlusions, and domain gaps all at once. Specifically, UnmaskFormer includes the crucial designs of Unmasking Attention (UA) and Amodal-oriented Mix (AoMix). Our method achieves state-of-the-art performance on the BlendPASS dataset, reaching a remarkable mAPQ of $26.58\%$ and mIoU of $43.66\%$. On public panoramic semantic segmentation datasets, \ie, SynPASS and DensePASS, our method outperforms previous methods and obtains $45.34\%$ and $48.08\%$ in mIoU, respectively. The fresh BlendPASS dataset and our source code are available at \url{https://github.com/yihong-97/OASS}.
\end{abstract}

\section{Introduction}
\label{sec:intro}
Panoramic imaging has advanced significantly~\cite{jiang2022annular,yu2023osrt}, allowing the capture of high-quality 360{\textdegree} images with minimalist optical systems~\cite{jiang2023minimalist} suitable for a wide variety of omnidirectional vision applications~\cite{ai2022deep_omnidirectional,gao2022review}. 
Concurrently, panoramic scene understanding has advanced in many areas, such as dense visual prediction~\cite{shen2022panoformer,tateno2018distortion,ling2023panoswin,yu2023panelnet},
holistic scene understanding~\cite{teng2023360bev,zhang2021deeppanocontext},
and panoramic scene segmentation~\cite{xu2019semantic,orhan2022semantic_outdoor,yang2020ds,hu2022distortion}. 
On the other side, amodal perception~\cite{nanay2018importance}, a fundamental aspect of human vision that forms the basis of our understanding and interpretation of the world, motivates occlusion-aware amodal prediction~\cite{mohan2022amodal_panoptic_segmentation, qi2019amodal_kins, ao2023image_survey} aimed at achieving recognition of an object and its complete spatial extent. These diverse research endeavors converge toward the common goal of achieving more comprehensive visual perception and understanding.
\begin{figure}[t]
        \begin{subfigure}[t]{0.58\linewidth}
            \includegraphics[width=0.99\linewidth]{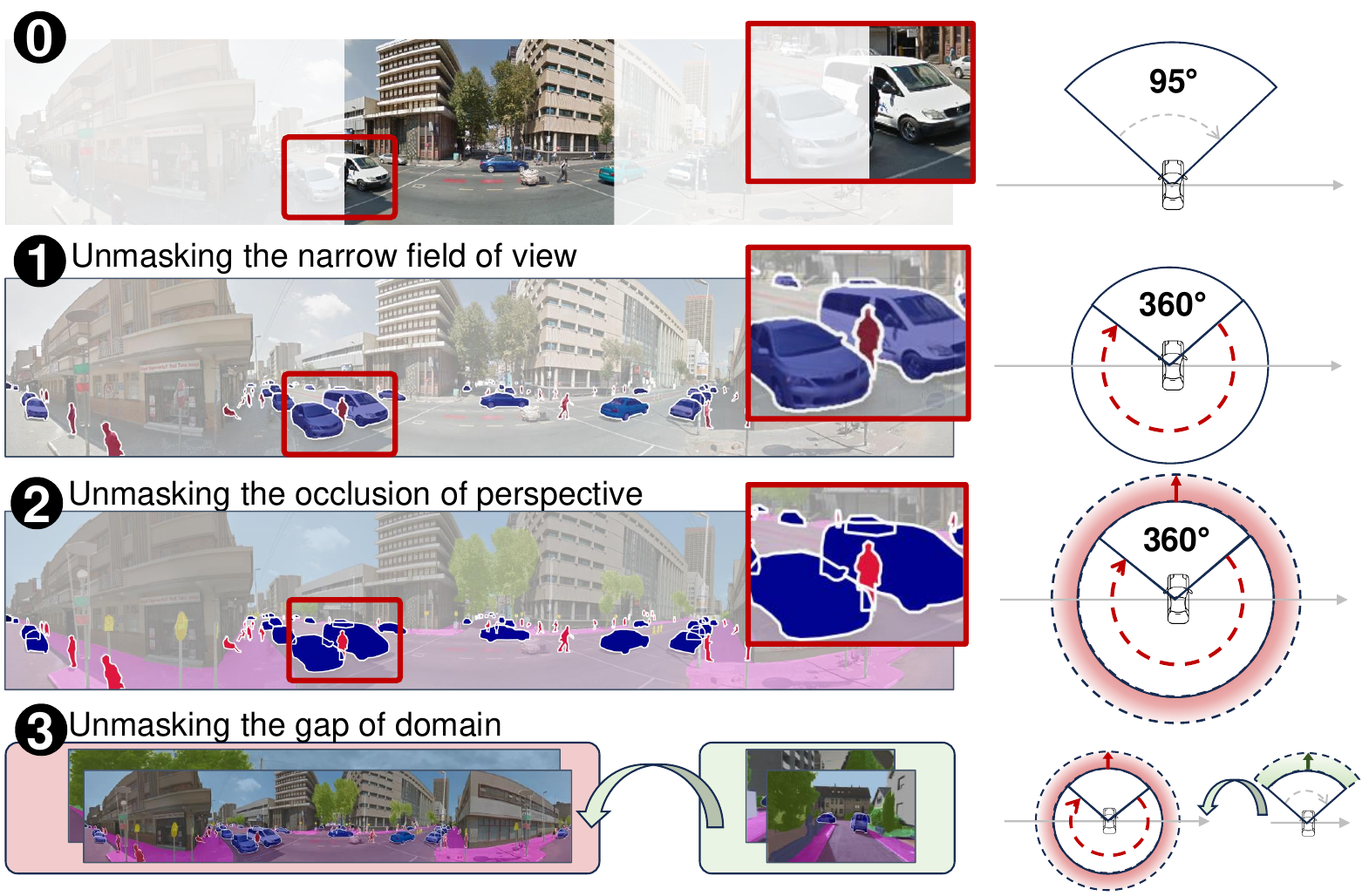}  
            \hfill
            \caption{\textbf{Occlusion-Aware Seamless Segmentation (OASS)} task involves three challenges: (1) unmasking the narrow field of view, (2) unmasking the occlusion of perspective, and (3) unmasking the gap of domain.}
            \label{fig:1a}
        \end{subfigure}
        \hspace {2ex}
        \begin{subfigure}[t]{0.39\linewidth}
            \includegraphics[width=0.99\linewidth]{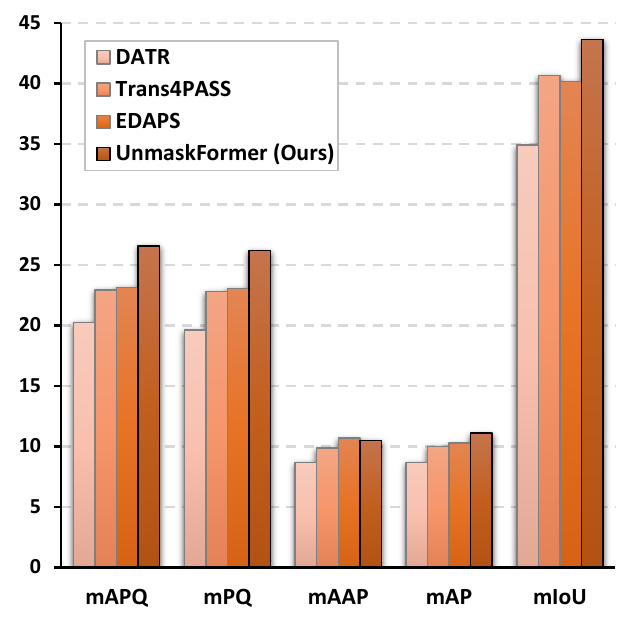}   
            \hfill
            \caption{\textbf{Results on our BlendPASS benchmark.}
            Domain adaptive panoptic and panoramic segmentation methods~\cite{saha2023edaps,zhang2022bending,zheng2023look_neighbor} are compared.}
            \label{fig:1b}
        \end{subfigure} 
	\caption{The overview, challenges, and comparison on the proposed Occlusion-Aware Seamless Segmentation (OASS) task. 
 }
    \label{fig1:oass}
\end{figure}

To unify the aforementioned scene understanding in a seamless form and advance a more comprehensive perception, we introduce \textbf{OASS (Occlusion-Aware Seamless Segmentation)}. 
As illustrated in \cref{fig1:oass}, the OASS task offers threefold benefits while posing three significant challenges: 
(1) Panoramic images offer a broader Field of View (FoV), unmasking the narrow FoV of the pinhole imagery, \eg, from 95{\textdegree} to 360{\textdegree}, as compared between \circled{0} and \circled{1} of \cref{fig:1a}. 
However, panoramas introduce severe distortions compared to pinhole images, which will lead to a significant performance degradation~\cite{zhang2022bending, zheng2023look_neighbor}. 
(2) Amodal prediction~\cite{ao2023image_survey}, in contrast to segmentation limited in visible areas~\cite{xie2021segformer, guo2022segnext}, can unmask the occlusion in the space perspective. 
For example, compared between \circled{1} and \circled{2} of \cref{fig:1a}, the occluded \textit{pedestrian} can be completely recognized by using amodal prediction. Predicting the complete mask of occluded objects, 
\ie, occlusion reasoning, 
is important to enhance the spatial-wise understanding capacity~\cite{chen2020banet,yuan2021robust_instance_segmentation,lazarow2020learning_instance_occlusion,ke2021deep_occlusion_aware}. 
(3) Unsupervised Domain Adaptation (UDA)~\cite{saha2023edaps, zhang2022UniDAPS} in \circled{3} is capable of addressing the need for expensive training data, unmasking domain gaps from label-rich pinhole to label-scare panoramic imagery.  

\begin{wrapfigure}{TR}{0.5\textwidth}
\centering
\includegraphics[width=0.5\textwidth]{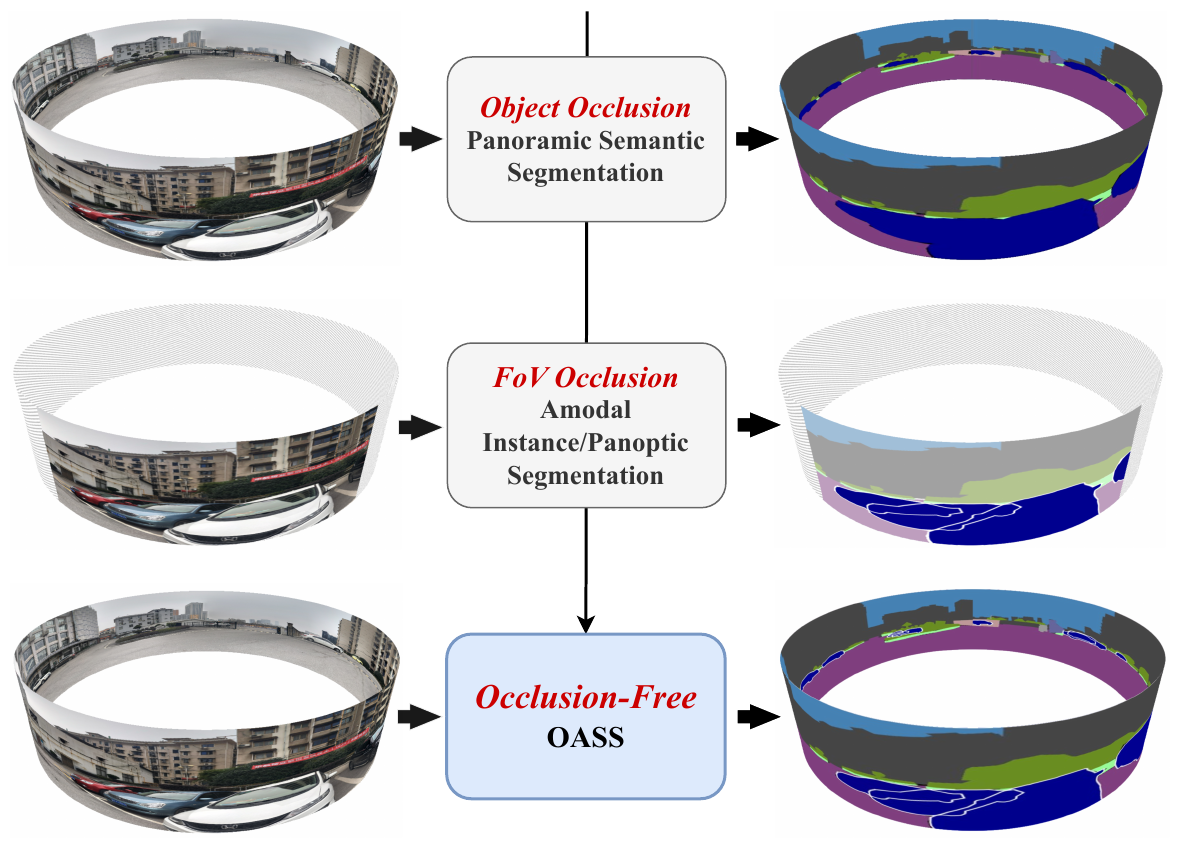}
\caption{OASS addresses \textit{object occlusion} and \textit{FoV occlusion} limitations in segmentation tasks.}
\label{fig:2-paradigm}
\end{wrapfigure}

Previous approaches~\cite{zhang2022bending, zheng2023look_neighbor, breitenstein2022amodal_cityscapes, qi2019amodal_kins, mohan2022amodal_panoptic_segmentation} are proposed to address the above challenges separately, resulting in sub-optimal and non-seamless solutions. 
For example, as the comparison depicted in \cref{fig:2-paradigm},
methods~\cite{zhang2022bending, zheng2023look_neighbor} proposed to address panoramic semantic segmentation cannot handle the object occlusion, whereas methods~\cite{qi2019amodal_kins, mohan2022amodal_panoptic_segmentation, breitenstein2022amodal_cityscapes} proposed for amodal segmentation cannot generalize well to panoramic imagery and the FoV occlusion remains unsolved. 
To address the challenges seamlessly, 
we propose a novel unmasking transformer framework called \textbf{UnmaskFormer}. 
The novelty is three-fold: (1) An \textit{Unmasking Attention (UA)} constructed by a self-attention and an enhanced pooling layer for occlusion prediction, is proposed to unmask the object occlusions within the whole UnmaskFormer framework. (2) We delve deeper into the design of the Deformable Patch Embedding (DPE)~\cite{zhang2022bending} and find an alternative yet better solution for addressing the image distortion of panoramas at different stages within the transformer model.  
(3) We propose an \textit{Amodal-oriented Mix (AoMix)} method that aims to seamlessly integrate the pinhole and panoramic domains, addressing the challenges posed by inconsistent cross-domain scenes. This method also enhances the model's capacity to reconstruct invisible regions of occluded objects, ultimately allowing it to unmask the occlusion of perspective within a scene.
Based on these crucial designs, our UnmaskFormer can better handle different panorama-based scene understanding tasks, especially solving object occlusion and image distortion.
\cref{fig:1b} shows that UnmaskFormer has striking performances across Semantic (mIoU), Instance (mAP), Amodal Instance (mAAP), Panoptic (mPQ), and Amodal Panoptic Segmentation (mAPQ) tasks by using one model.

To facilitate this evaluation, we spent a large effort to collect and manually annotate a dataset for \textbf{Blending Panoramic Amodal Seamless Segmentation (BlendPASS)}.
There are $2,000$ panoramic images with 360{\textdegree} FoV and a $2048{\times}400$ resolution, captured in street scenes.
BlendPASS facilitates the learning optimization of segmentation adaptation in an unsupervised fashion, which unfolds as efficient for dense prediction in panoramic imagery.
To establish the evaluation benchmark for the OASS task, $100$ panoramic images have been manually annotated precisely at the pixel level.

Extensive experiments are conducted on our proposed BlendPASS dataset and other panoramic datasets. Our UnmaskFormer achieves state-of-the-art performance on BlendPASS, reaching a remarkable mAPQ of $26.58\%$ and mIoU of $43.66\%$. On the panoramic semantic segmentation datasets SynPASS~\cite{zhang2022behind} and DensePASS~\cite{ma2021densepass}, our method outperforms the previous best method and obtains $45.34\%$ and $48.08\%$ in mIoU, respectively. 
The significant improvements and results prove the effectiveness of the proposed UnmaskFormer framework in addressing the panorama-based scene understanding.

In this work, we propose contributions as follows:
\begin{itemize}
    \item  We introduce a new panorama-based segmentation task, \ie, \textbf{OASS}, aiming at unmasking the narrow field of view, unmasking the occlusion of perspective, and unmasking the gap of domain in a seamless manner. 
    \item  To address the OASS task, we propose an \textbf{UnmaskFormer} framework with distortion- and occlusion-aware designs in a transformer-based architecture.
    \item  An \emph{Amodal-oriented Mix (AoMix)} method is tailored for improving unsupervised domain adaptation and overcoming the challenges of diverse occlusions in bridging the gap between pinhole and panoramic domains.
    \item  A new panoramic dataset \emph{BlendPASS} is created and manually annotated for benchmarking the blending of panoramic amodal seamless segmentation. 
\end{itemize}

\section{Related Work}
\label{related_work}
\noindent\textbf{Domain adaptive panoramic segmentation.}
Segmentation on wide-FoV fisheye images~\cite{deng2017cnn,shi2023fishdreamer,ye2020universal,sekkat2020omniscape,yogamani2019woodscape} and panoramic images~\cite{guttikonda2023single_spherical,li2023sgat4pass,yang2019pass,yang2021context,zheng2023complementary_bidirectional} enables holistic understanding of 360{\textdegree} surroundings~\cite{yang2019can}.
To address the scarcity of annotations, researchers have revisited wide-FoV segmentation from the perspective of UDA~\cite{kim2022pasts,jang2022dada_uda,yang2021capturing,yang2020omnisupervised,shi2022unsupervised} by leveraging rich training sources from conventional narrow-FoV data.
A variety of adaptation methods including self-training~\cite{hoyer2022hrda,hoyer2023mic,zou2019crst,hoyer2022daformer,wang2020differential,yue2021prototypical} and adversarial training~\cite{ma2021densepass,luo2019taking_clan} methods are studied for panoramic segmentation.
In this line, P2PDA~\cite{zhang2021transfer} employs attention-regulated uncertainty-aware adaptation.
Trans4PASS~\cite{zhang2022bending,zhang2022behind} introduces distortion-aware transformers and mutual prototypical adaptation with SAM~\cite{kirillov2023segment}.
DPPASS~\cite{zheng2023both_style_distortion} presents a dual-path solution to overcome style and format gaps. 
DATR~\cite{zheng2023look_neighbor} captures neighboring distributions without any geometric constraints. 
Moreover, panoramic panoptic segmentation~\cite{fu2023panopticnerf,jaus2023panoramic_insights,mei2022waymo} is investigated
by using techniques like contrastive learning for harvesting generalization benefits~\cite{jaus2021panoramic_towards}.
In this work, we aim to enable occlusion-informed understanding with both FoV-wise and spatially amodal reasoning of urban scenes.

\noindent\textbf{Amodal scene segmentation.}
Amodal instance segmentation is a derivative task of instance segmentation~\cite{hariharan2014simultaneous,dai2015convolutional,back2022unseen_occlusion,sun2022amodal_bayesian,tran2022aisformer,li2023muva,fan2023rethinking,liu2024blade}, to predict visible regions of objects and their occluded regions simultaneously. 
Li~\etal~\cite{li2016amodal_instance_segmentation} introduce the concept of amodal instance segmentation, achieved through iterative regression.
Qi~\etal~\cite{qi2019amodal_kins} derive the amodal version of KITTI~\cite{geiger2013vision}, introducing an independent occlusion classification branch. 
SLN~\cite{zhang2019learning_distance} incorporates a semantics-aware distance map to implement amodal segmentation.
ORCNN~\cite{follmann2019learning_see_invisible} calculates visible masks to infer occlusion masks. 
Further, shape and contour priors are frequently leveraged for refining amodal segmentation~\cite{gao2023coarse,chen2023amodal_expansion,li2023gin,xiao2021amodal_segmentation_prior,li20222d_3d_prior}.
Aside from amodal instance segmentation, amodal semantic understanding~\cite{zhu2017semantic_amodal_segmentation,hu2019sail,breitenstein2022amodal_cityscapes} is an important direction.
Along this line, \cite{mohan2022amodal_panoptic_segmentation,mohan2022perceiving} propose a fusion of semantic and amodal instance segmentation. 
Sekkat~\etal~\cite{sekkat2023amodalsynthdrive} uses the CARLA simulator~\cite{dosovitskiy2017carla} to create a virtual multi-task amodal perception dataset. 
Different from these works, we intend to achieve occlusion-aware seamless segmentation, which breaks the occlusion limits in terms of both field-of-view and in-field object occlusions.

\section{Established Benchmark}
\subsection{Overview of the Benchmark}
\label{3.1}
In this work, we establish a novel benchmark specifically designed for OASS.
In particular, we address OASS from the perspective of UDA by adapting from the label-rich pinhole domain to the label-scarce panoramic domain.
360{\textdegree} panoramas have a wide FoV and many small objects in the images, which exaggerate the costs of creating 
pixel-wise annotations in unconstrained surroundings.
Our benchmark overcomes the scarcity of amodal segmentation testbeds in panoramic imagery, 
while addressing three aforementioned challenges:
\circled{1} \textit{how to address the severe distortions of panoramas when unmasking the narrow field of view},
\circled{2} \textit{how to predict the full segmentation of objects when unmasking the occlusion of perspective},
and \circled{3} \textit{how to facilitate optimization of segmentation adaptation when unmasking the gap of domain}.

Our objective with this benchmark is to provide a comprehensive evaluation of methods capable of performing both FoV-wise and spatially occlusion-aware seamless segmentation. 
We extend the metrics proposed by~\cite{kirillov2019panoptic} to the full regions incorporating pixels of the occluded objects.
The benchmark metrics cover three aspects: 
Intersection over Union (IoU) for semantic segmentation, Average Precision (AP) and Panoptic Quality (PQ) for instance and panoptic segmentation, Amodal Average Precision (AAP) and Amodal Panoptic Quality (APQ) for amodal instance and amodal panoptic segmentation.
\begin{figure}[h]
    \centering
    \includegraphics[width=1\linewidth]{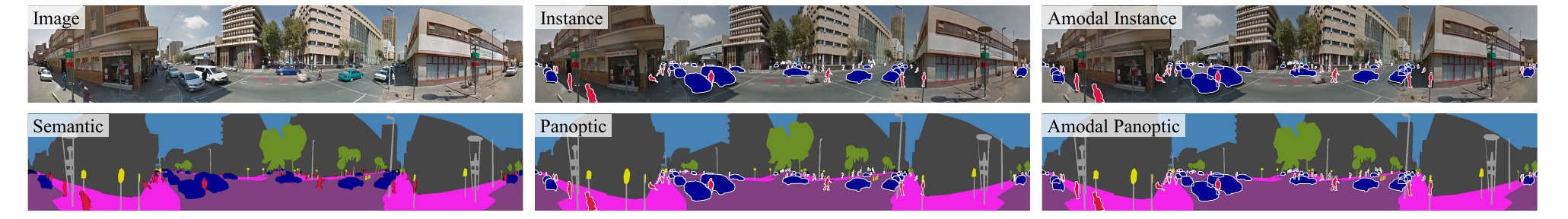}
   
    \label{BlenPAS_a}
 
  \caption{\textbf{The established BlendPASS dataset}. We provide pixel-level labels for five segmentation tasks related to OASS on the validation set. Zoom in for a better view.}
  \label{BlenPAS}
\end{figure}

\subsection{BlendPASS}
We introduce a novel dataset for \textit{Blending Panoramic Amodal Seamless Segmentation} (BlendPASS) tailored for OASS.
BlendPASS comprises an unlabeled training set of $2,000$ panoramic images for optimizing domain adaptation and a labeled test set of $100$ panoramic images.
These images are captured from panoramic cameras in driving scenes,
all at a resolution of $2048{\times}400$ pixels. 
As depicted in \cref{BlenPAS_a}, we have provided pixel-wise annotations for five distinct visual segmentation tasks, which greatly extends the semantic labels from DensePASS~\cite{ma2021densepass}. 
Specifically, we have further annotated the instance and amodal instance labels.
These labels cover $19$ categories that align with the Cityscapes~\cite{cityscapes} and are further categorized into \textit{Stuff} (\textit{road, sidewalk, building, wall, fence, pole, light, sign, vegetation, terrain, and sky}) and \textit{Thing} (\textit{person, rider, car, truck, bus, train, motorcycle, and bicycle}). 
To ensure the precision of annotations and the rational handling of occluded object parts, we meticulously conduct manual annotation for the test set, with three annotators following a cross-checking process. Finally, $2,960$ objects are annotated in the \textit{Thing} class, with $43\%$ of these objects exhibiting occlusion. More details can be found in the supplementary.

\subsection{KITTI360-APS$\rightarrow$BlendPASS}
For the labeled source in our UDA benchmark for OASS, we employ the available KITTI360-APS dataset~\cite{mohan2022amodal_panoptic_segmentation} designed for amodal panoptic segmentation. 
This dataset is an extension of KITTI360~\cite{Liao2022PAMI} and includes annotations for inmodal/amodal instance and panoptic segmentation.
The images in KITTI360-APS are captured using pinhole cameras from $9$ cities, with a resolution of $1408{\times}376$. 
After our careful examination, a total of $12,320$ annotated images are accessed. 
These annotations encompass $10$ \textit{Stuff} (\textit{road, sidewalk, building, wall, fence, pole, traffic sign, vegetation, terrain, and sky}) classes and $7$ \textit{Thing} (\textit{car, pedestrians, cyclists, two-wheeler, van, truck, and other vehicles}) classes. 
In our benchmark, we further process the annotations from BlendPASS to ensure class alignment with KITTI360-APS.

\begin{figure}[h]
    \begin{subfigure}[t]{0.49\linewidth}
    \centering
    \includegraphics[width=\linewidth]{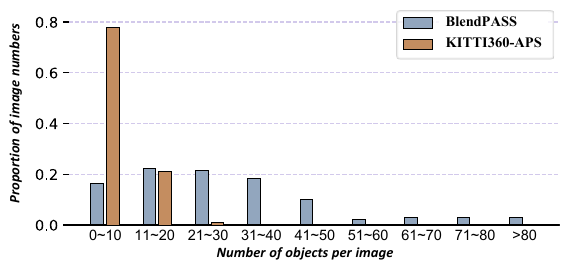}
    \caption{Statistics of object counts per image.}
    \hfill
    \label{fig_dataUDA_a}
  \end{subfigure}   
  \hfill
  \begin{subfigure}[t]{0.49\linewidth}
    \centering
    \includegraphics[width=\linewidth]{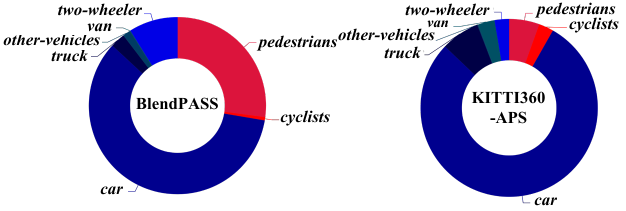}
    \caption{Class distribution of the dataset.}
    \label{fig_dataUDA_b}
  \end{subfigure}  
    \caption{Analysis of test set (BlendPASS) and the training set (KITTI360-APS).}
    \label{fig_dataUDA}
\end{figure}

For KITTI360-APS${\rightarrow}$BlendPASS under OASS, besides the challenges mentioned in \cref{3.1}, as illustrated in \cref{fig_dataUDA}, the source-target domains exhibit significant differences in terms of the number of objects per image and the class distribution.
These differences present a greater challenge to UDA models.

\section{Methodology}
\label{methodology}

\subsection{UnmaskFormer}
\begin{figure*}
    \centering
    \includegraphics[width=\textwidth]{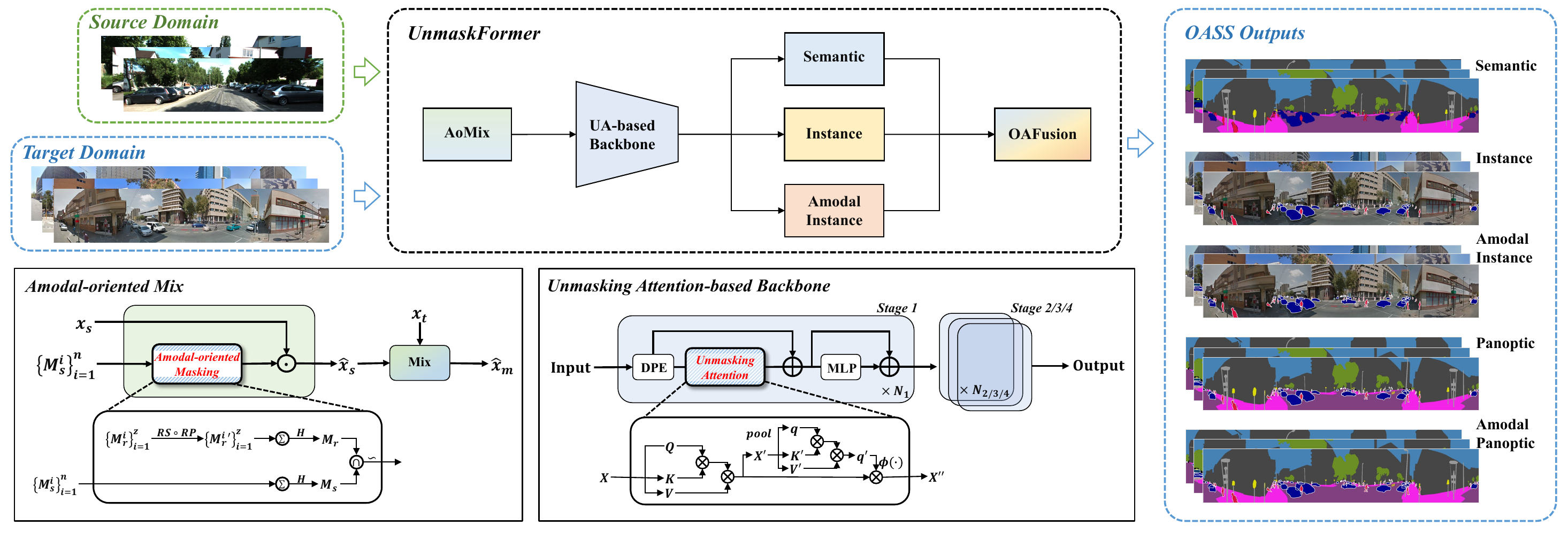}
    \caption{\textbf{Overview of the proposed UnmaskFormer framework.} The UA-based backbone consists of the DPE~\cite{zhang2022bending} and the proposed UA, addressing image distortion and object occlusion in panoramas at once. The AoMix is designed to process input images $x_s, x_t$, seamlessly integrating pinhole and panoramic domains while enhancing the model's capacity to reconstruct invisible regions of the occluded objects. 
    } 
    \label{fig_framework}
\end{figure*}

\noindent\textbf{Architecture.}
For the OASS task, the model simultaneously addresses objectives including semantic, instance, amodal instance, panoptic, and amodal panoptic segmentation. 
Similarly to panoptic segmentation, as illustrated in \cref{fig_framework}, we decompose the base model $\mathcal{F}$ into a feature extractor and three branches, \ie, semantic, instance, and amodal instance, to accomplish OASS goals.
The semantic branch predicts the per-pixel semantic category of the input image,
while the instance and amodal instance branches output class-agnostic object localization predictions.
It is noteworthy that, while the instance branch has both top-down~\cite{saha2023edaps} and bottom-up~\cite{zhang2022UniDAPS} decoders, the contour-based bottom-up decoder is impractical for the amodal instance branch in the context of OASS. 
This limitation arises from situations where a single pixel is associated with multiple objects simultaneously. 
In contrast, the proposals-based top-down decoder typically follows Mask R-CNN~\cite{he2017mask}, employing Region Proposal Networks (RPN) to predict candidate objects. 
This methodology facilitates simultaneous and interference-free segmentation of multiple objects. 
Hence, in UnmaskFormer, we adopt the top-down decoder for the instance branch and the occlusion-aware amodal instance branch.

For the final output of the UnmaskFormer, 
we construct an Occlusion-Aware Fusion (OAFusion) module to process outputs from three branches, generating five segmentation maps at once for OASS.
The semantic segmentation is directly predicted by the semantic branch. 
In instance and amodal instance segmentation, the semantic label of class-agnostic instance or amodal instance mask is determined by the majority-voting rule of the predictions from the semantic branch.
For amodal segmentation, only regions where the current object does not overlap with other objects are considered. 
The panoptic and amodal panoptic segmentation are generated by fusing the semantic segmentation with instance and amodal instance segmentation, respectively.

\noindent\textbf{UA-based backbone.}
One of challenges in OASS is to address both image distortion and object occlusion at the same time. For this end, we construct a UA-based backbone by rearranging previous Deformable Patch Embedding (DPE) and adding an effective pooling layer, as shown in \cref{fig_framework}. 
Specifically, for an input target image or features $\boldsymbol{X}{\in}\mathbb{R}^{H{\times}W{\times}C}$ between different stages, the original DPE calculates the adaptive offsets and performs the patchifying process. 
Contrary to inserting DPE in the early stage~\cite{zhang2022bending}, we perform a new interleaving arrangement of DPE layers in Stages 2 and 4, which can provide reinforced distortion-aware modeling for the whole UnmaskFormer framework. More analysis will be presented in Sec.~\ref{sec:ablation_study}. 

Apart from the distortion-aware design, we design the UA block by combining the self-attention layer and the enhanced pooling layer. 
As shown in \cref{fig_framework}, after the self-attention layer, the feature $\boldsymbol{X'}{\in}\mathbb{R}^{H{\times}W{\times}C}$ is forwarded to calculate the pooling query {$\boldsymbol{q}$}, the key $\boldsymbol{K'}$ and the value $\boldsymbol{V'}$, where $\boldsymbol{q}{=}{GAP(\boldsymbol{X'})}$ and $GAP(\cdot)$ is the global average pooling operator.
After having the self-attended pooling feature  {$\boldsymbol{q'}$}${\in}\mathbb{R}^{1{\times}1{\times}C}$, a \texttt{sigmoid} function $\phi(\cdot)$ is applied to calculate the occlusion-aware mask $\phi({{\boldsymbol{q'}}})$. Then, the pool-attended occlusion-aware mask $\phi({{\boldsymbol{q'}}})$ is used to perform a dot product operation with feature $\boldsymbol{X'}$ to obtain the occlusion-aware feature {$\boldsymbol{X''}$}${\in}\mathbb{R}^{H{\times}W{\times}C}$. After the UA block, the occlusion-aware feature {$\boldsymbol{X''}$} is further forwarded to an MLP layer as self-attention blocks~\cite{xie2021segformer, zhang2022bending}. 

The overall UA-based backbone is constructed as the same four-stage architecture as the common segmentation methods~\cite{zhang2022bending, wang2021pvt, xie2021segformer}.

\subsection{Cross-Domain Adaptation}
\noindent\textbf{UDA strategy.}
In the OASS setting, the model is assigned to adapt from a labeled source domain $\mathcal{X}_s=\{{x_s, y_s}\}$ to an unlabeled target domain $\mathcal{X}_t=\{{x_t}\}$ for multiple segmentation tasks, which $x_s\in \mathbb{R}^{H^s\times W^s\times3}, x_t \in \mathbb{R}^{H^t\times W^t\times3}$ are images from pinhole and panoramic cameras respectively.
To address the domain gap arising from dissimilar data distributions between the source pinhole and target panoramic domains, a comprehensive loss $\mathcal{L}_{total} = \mathcal{L}_S+\mathcal{L}_T$ is utilized for training the overall network $\mathcal{F}$,
where $\mathcal{L}_S$ and $\mathcal{L}_T$ represent the source domain supervision loss and the target domain adaptation loss, respectively.
$\mathcal{L}_S$ is utilized for supervision to learn fundamental segmentation capacity through labeled source samples $\{x_s, y_s\}$, whereas the $\mathcal{L}_T$ leverages unlabeled target images $x_t$ to enhance segmentation in the target panoramic scenario. $\mathcal{L}_S$ includes semantic loss, instance loss, and amodal instance loss, each originating from respective branches. 
The semantic branch employs cross-entropy loss to assign each pixel to its respective category. 
As for the instance and amodal instance branches, we follow the conventional approach from Mask R-CNN~\cite{he2017mask}, employing supervision using proposals-based bounding boxes and instance-level masks.

To tackle the challenge posed by the absence of labels in the target domain, we employ a self-training strategy to facilitate the model's adaptation from the source to the target domain. Self-training methods \cite{hoyer2022hrda,hoyer2023mic,hoyer2022daformer} typically utilize the target predictions $p_t$ as pseudo-labels $\hat{y}_t$ for training. However, due to the noisy predictions in the early stage, resulting in low-confidence pseudo-labels, we adopt a confidence estimation mechanism that assigns weights $\omega$ of pseudo-labels-based self-training loss $\mathcal{L}_{T}$.
Additionally, we incorporate the Mean-Teacher framework widely adopted in UDA~\cite{hoyer2022daformer, tranheden2021dacs, saha2023edaps} to further enhance the quality of the pseudo-labels. Thus, $\mathcal{L}_{T}$ can be expressed as follows:
\begin{equation}
    \mathcal{L}_{T} = - {\omega}\sum\nolimits_{h,w,c}{p_t^{(h,w,c)}\log{\hat{y}_t^{(h,w,c)}}},
\end{equation}
where 
\begin{equation}
    {\omega} = \frac{1}{H^t\cdot W^t} \sum\nolimits_{h,w}{(\max\limits_{c}{\mathcal{T}(x_t)}^{(h,w,c)}>\tau)},
\end{equation}
\begin{equation}
    \hat{y}_t^{(h,w)}=onehot(\arg\max\limits_{c}{\mathcal{T}(x_t)}^{(h,w,c)}).
\end{equation}
The parameters $\theta'$ of the teacher model $\mathcal{T}$ are updated using the parameters $\theta$ of the student model $\mathcal{F}$ by the EMA at each iteration.

\noindent\textbf{Amodal-oriented Mix (AoMix).}
To address the challenges of diverse occlusion cases and inconsistent cross-domain scenes, we propose a new method AoMix, which processes images from two domains alongside self-training.
Initially, we reconstruct the masked source images $\hat{x}_s$ based on a random source amodal annotations $\{M_r^i\}_{i=1}^z$. 
Subsequently, we follow the widely-used class-mix strategy~\cite{tranheden2021dacs,hoyer2022daformer} in UDA segmentation tasks to generate masked mixed images ${\hat{x}_m}$, incorporating information from both the source and target domains. 

Specifically, we randomly sample an amodal instance mask sequence $\{M_r^{(i)}\}_{i=1}^z$ of an image from the current batch, and subject it to random scaling $RS(\cdot)$ and peripheral random padding $RP(\cdot)$ to generate a new amodal instance mask sequence $\{{M_r^{(i)}}'\}_{i=1}^z$ with the object of various positions and sizes. 
Then, a random binary mask $M_r$ containing multiple objects is produced by summing these masks $\{{M_r^{(i)}}'\}_{i=1}^z$. 
This operation can be expressed as follows:
\begin{equation}
    M_r = H(\sum\nolimits_i{RP(RS(M_r^{(i)}))}),
\end{equation}
where $H(\cdot)$ denoted as a step function.
For a source image $x_s$, we sum the corresponding amodal instance masks $\{M_s^i\}_{i=1}^n$ to obtain a binary mask $M_s$ of \textit{Thing} region. Using the random binary mask $M_r$ and the \textit{Thing} region binary mask $M_s$, we fill the source image to reconstruct the masked source image $\hat{x}_s = (1-M_r\cap M_s) \odot x_s$.
Amodal-oriented masked image modeling aims to enhance the model's ability to reconstruct object regions obscured by realistic object shapes, enabling it to learn information about invisible parts. 
To adapt to unlabeled target scenes, we randomly sample half of the semantic classes from a source image and paste the corresponding pixels of the masked source image $\hat{x}_s$ onto a target image ${x}_t$ based on the source semantic map, creating a new masked mixed image $\hat{x}_m$.
This approach effectively transfers the model's capability tailored for distortion- and occlusion-aware to the target panoramic domain.

\section{Experiments}
\label{experiments}
\subsection{Experiment Setups}
Following DAFomer~\cite{hoyer2022daformer}, we train UnmaskFomer using AdamW~\cite{loshchilov2017decoupled}. 
The learning rate is set to $6{\times}10^{-5}$, with a weight decay of $0.01$ and linear warmup for $1.5k$ iterations followed by polynomial decay. 
The model is trained on a batch size of $4$ with a crop size of $376{\times}376$ for $40k$ iterations. 
More details can be found in the supplementary.

\begin{table}[htbp]
\centering
\caption{
\textbf{Occlusion-Aware Scene Segmentation} results on the KITTI360-APS $\rightarrow$ BlendPASS benchmark. For Semantic Segmentation (\textbf{SS}), per-class results are reported as IoU, and the metric is the mIoU. For Amodal Panoptic Segmentation (\textbf{APS}), per-class results are reported as APQ for full regions, and the Metric is the mAPQ.}
\resizebox{\linewidth}{!}{
\begin{tabular}{c  |l|cccccccccccccccccc|c}
\toprule 
 \multicolumn{1}{c}{Task}&\multicolumn{1}{l}{UDA Method} & \rots{road} & \rots{sidewalk} & \rots{building} & \rots{wall} & \rots{fence} & \rots{pole} & \rots{traffic-light} & \rots{traffic-sign} & \rots{vegetation} & \rots{terrain} & \rots{sky} & \rots{pedestrians} & \rots{cyclists} & \rots{car} & \rots{truck} & \rots{other-vehicles} & \rots{van} & \rots{two-wheeler} &  Metric \\
\hline\hline
\multirow{6}[0]{*}{\textbf{SS}}
&DATR~\cite{zheng2023look_neighbor} & 71.87&27.24&70.59&22.76&35.98&23.91&00.00&04.52&77.06&37.14&80.06&51.20&02.24&70.15&08.94&06.00&11.02&27.79&34.91\\
&Trans4PASS~\cite{zhang2022bending} &72.79&33.80&78.38&33.53&37.07&26.56&05.31&05.35&77.43&37.91&84.73&57.48&04.51&76.58&19.04&17.03&12.66&51.70&40.66\\
&UniDAPS \cite{zhang2022UniDAPS} &72.27&29.20&75.78&33.91&38.92&25.94&11.37&07.51&77.61&37.40&81.40&47.56&02.95&75.59&21.29&03.39&01.31&48.85&38.46\\
&EDAPS~\cite{saha2023edaps} &74.41&35.91&76.98&36.45&40.35&28.02&16.08&05.13&78.10&39.50&82.28&55.45&03.25&74.38&06.72&14.09&05.18&50.87&40.17\\
&Source-Only&72.97&29.14&82.04&31.19&31.37&17.98&00.00&15.97&74.13&33.74&88.66&50.24&04.02&80.42&18.02&11.04&04.57&50.17&38.65\\
&\cellcolor[gray]{.9}UnmaskFormer
&\cellcolor[gray]{.9}76.54&\cellcolor[gray]{.9}37.82&\cellcolor[gray]{.9}77.06&\cellcolor[gray]{.9}34.71&\cellcolor[gray]{.9}44.05&\cellcolor[gray]{.9}28.27&\cellcolor[gray]{.9}17.80&\cellcolor[gray]{.9}02.76&\cellcolor[gray]{.9}78.70&\cellcolor[gray]{.9}41.68&\cellcolor[gray]{.9}84.98&\cellcolor[gray]{.9}57.34&\cellcolor[gray]{.9}06.01&\cellcolor[gray]{.9}80.60&\cellcolor[gray]{.9}23.47&\cellcolor[gray]{.9}21.67&\cellcolor[gray]{.9}18.80&\cellcolor[gray]{.9}53.56&\cellcolor[gray]{.9}\textbf{43.66}\\

\midrule

\multirow{5}[0]{*}{\textbf{APS}}
& DATR~\cite{zheng2023look_neighbor} &51.82&09.15&59.90&11.93&11.98&01.97&00.00&03.91&64.61&14.05&70.40&11.09&00.00&39.30&00.00&03.16&10.07&01.30&20.26\\
&Trans4PASS~\cite{zhang2022bending} &53.91&14.12&69.39&19.15&11.77&03.77&00.00&05.15&67.63&16.02&77.41&15.30&04.24&41.06&06.58&00.00&00.00&07.35&22.94\\
&EDAPS~\cite{saha2023edaps} &54.88&17.04&66.86&18.75&14.47&05.75&04.04&04.64&68.20&16.04&72.76&19.01&00.00&36.73&05.77&04.38&00.00&07.20&23.14 \\
&Source-Only&57.11&14.21&73.58&15.49&07.59&00.67&00.00&10.40&58.30&12.39&83.14&15.23&00.00&40.31&03.83&00.00&00.00&06.08&22.13\\

&\cellcolor[gray]{.9}UnmaskFormer
&\cellcolor[gray]{.9}61.84&\cellcolor[gray]{.9}24.72&\cellcolor[gray]{.9}66.81&\cellcolor[gray]{.9}20.77&\cellcolor[gray]{.9}15.80&\cellcolor[gray]{.9}05.25&\cellcolor[gray]{.9}04.29&\cellcolor[gray]{.9}03.26&\cellcolor[gray]{.9}69.02&\cellcolor[gray]{.9}18.35&\cellcolor[gray]{.9}79.44&\cellcolor[gray]{.9}20.48&\cellcolor[gray]{.9}03.14&\cellcolor[gray]{.9}44.56&\cellcolor[gray]{.9}12.81&\cellcolor[gray]{.9}11.25&\cellcolor[gray]{.9}00.00&\cellcolor[gray]{.9}16.64&\cellcolor[gray]{.9}\textbf{26.58}\\

\bottomrule
\end{tabular}
}
\label{table:uda_sota_k2d}
\end{table}
\subsection{Results of OASS on KITTI360-APS$\rightarrow$BlendPASS}
We deliver the OASS results on the KITTI360-APS$\rightarrow$BlendPASS benchmark in \cref{table:uda_sota_k2d}. 
Representative UDA panoptic segmentation methods UniDAPS~\cite{zhang2022UniDAPS} and EDAPS~\cite{saha2023edaps} with our amodal extensions are benchmarked.
Further, SOTA panoramic semantic segmentation models Trans4PASS~\cite{zhang2022bending} and DATR~\cite{zheng2023look_neighbor} with distortion-adaptive capacities are compared.
It is worth noting that UniDAPS does not support occlusion-aware reasoning with a contour-based decoder.

\begin{figure}[t]
    \centering
    {
    \newcolumntype{P}[1]{>{\centering\arraybackslash}p{#1}}
    \renewcommand{\arraystretch}{1}
    \resizebox{0.93\linewidth}{!}{
    \begin{tabular}{@{}*{12}{P{0.115\columnwidth}}@{}}
    \textit{Stuff:}
    &{\cellcolor[rgb]{0.5,0.25,0.5}}\textcolor{white}{road}
    &{\cellcolor[rgb]{0.957,0.137,0.91}}sidew.
    &{\cellcolor[rgb]{0.275,0.275,0.275}}\textcolor{white}{build.}
    &{\cellcolor[rgb]{0.4,0.4,0.612}}\textcolor{white}{wall}
    &{\cellcolor[rgb]{0.745,0.6,0.6}}fence
    &{\cellcolor[rgb]{0.6,0.6,0.6}}pole
    &{\cellcolor[rgb]{0.98,0.667,0.118}}tr.light
    &{\cellcolor[rgb]{0.863,0.863,0}}tr.sign
    &{\cellcolor[rgb]{0.42,0.557,0.137}}veget.
    &{\cellcolor[rgb]{0.596,0.984,0.596}}terrain
    &{\cellcolor[rgb]{0.275,0.51,0.706}}sky\\
    
    \textit{Thing:}
    &{{\cellcolor[rgb]{0.863,0.078,0.235}}\textcolor{white}{pedes.}}
    &{\cellcolor[rgb]{1,0,0}}\textcolor{black}{cyclists}
    &{\cellcolor[rgb]{0,0,0.557}}\textcolor{white}{car}
    &{\cellcolor[rgb]{0,0,0.275}}\textcolor{white}{truck}
    &{{\cellcolor[rgb]{0,0.235,0.392}}\textcolor{white}{ot.veh.}}
    &{\cellcolor[rgb]{0,0.314,0.392}}\textcolor{white}{van}
    &{{\cellcolor[rgb]{0,0,0.902}}\textcolor{white}{tw.whe.}}\\
    \end{tabular}
    }
    }
    \includegraphics[width=\linewidth]{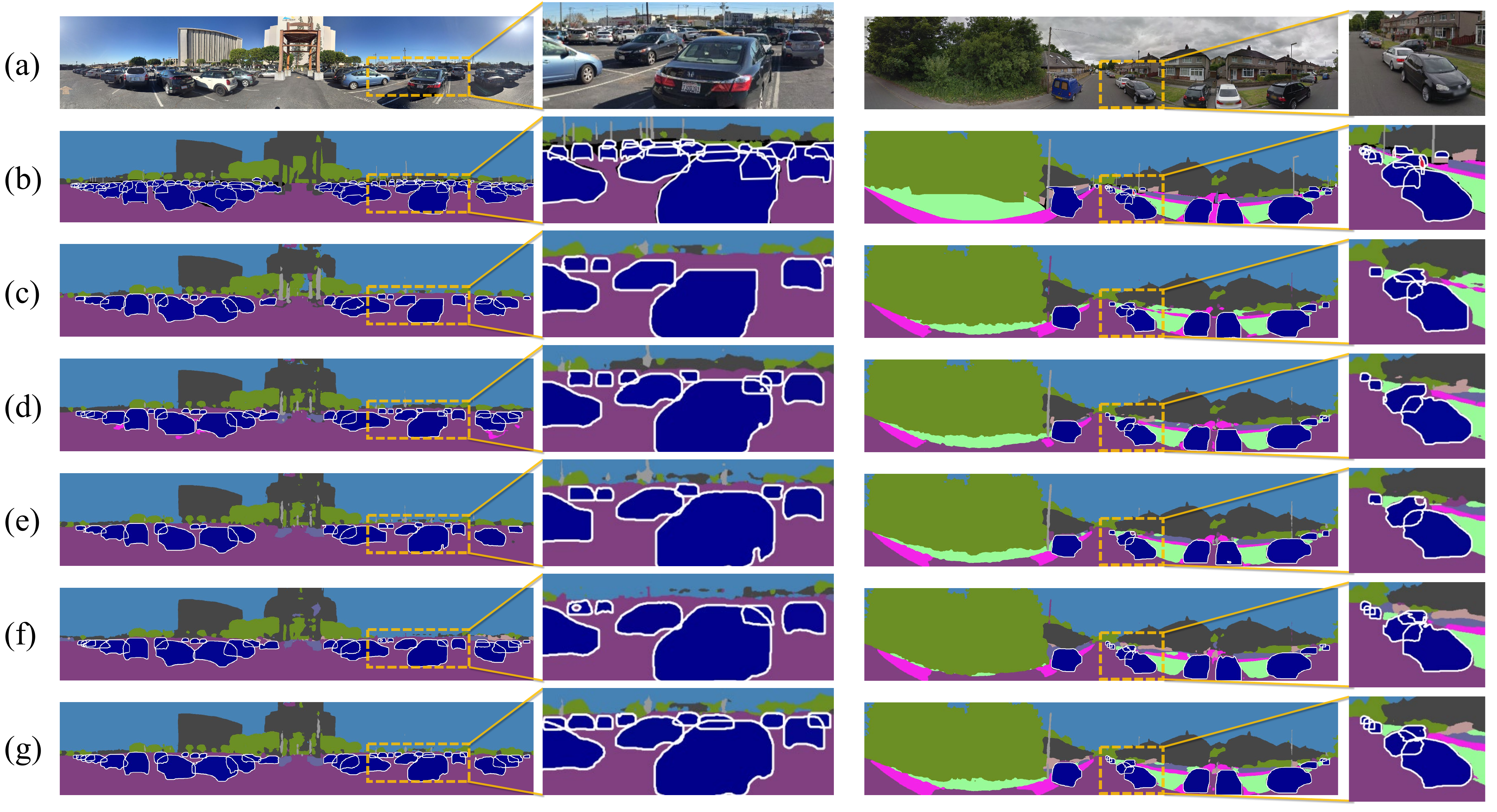}
    \caption{\textbf{Visualization results of OASS.} 
    From top to bottom are
    (a) Image, (b) Ground truth, (c) DATR~\cite{zheng2023look_neighbor}, (d) Trans4PASS~\cite{zhang2022bending}, (e) EDAPS~\cite{saha2023edaps}, (f) Source-only, and (g) UnmaskFormer (ours).}
    \label{fig_qualitative}
\end{figure}

As shown in \cref{table:uda_sota_k2d}, compared with the best-performing UDA panoptic segmentation model EDAPS~\cite{saha2023edaps}, UnmaskFormer outstrips it by clear margins of $3.49\%$ in mIoU and $3.44\%$ in mAPQ. 
Compared to the panoramic segmentation model Trans4PASS~\cite{zhang2022bending}, our method also yields significant gains.
Delving into the OASS results in semantic- and amodal panoptic segmentation, all benchmarked methods suffer from accurately predicting the full segmentation in the cases considering occlusions, in particular on small objects, evidenced by the unsatisfactory scores in class-wise APQ.
Yet, UnmaskFormer harvests great improvements in contrast to the Source-Only model, \eg, on safety-critical \emph{pedestrians}, \emph{truck}, and \emph{two-wheeler}.
Finally, under the challenging occlusion scenarios in unstructured surroundings, UnmaskFormer clearly stands out and leads to SOTA OASS scores of $43.66\%$ in mIoU and $26.58\%$ in mAPQ. 
In \cref{fig_qualitative}, we visualize OASS results produced by our UnmaskFormer and SOTA methods DATR~\cite{zheng2023look_neighbor}, Trans4PASS~\cite{zhang2022bending}, EDAPS~\cite{saha2023edaps}.
Compared to them, UnmaskFormer excels in segmenting occluded objects and realistically reconstructing vehicle shapes thanks to our UA-based backbone and AoMix design.

\newcommand*\rotsf{\multicolumn{1}{R{50}{1em}}}
\begingroup
\begin{table}[t!]
\centering
\caption{
\textbf{Instance-level Segmentation} results on the KITTI360-APS$\rightarrow$BlendPASS benchmark. 
Per-class results are reported as AP for visible regions and AAP for full regions. 
The Metrics are mAP and mAAP. 
``A'' denotes Amodal.
}
\resizebox{\linewidth}{!}{
\begin{tabular}{c|c|ccccccc |c }
\toprule 
UDA Method& A & {pedestrians} & {cyclists} & {car} & {truck} & {other-vehicles} & {van} & {two-wheeler} & Metric \\
\hline\hline
\multirow{2}[0]{*}{DATR~\cite{zheng2023look_neighbor}}& &14.21&00.00&31.15&07.55&03.73&00.39&03.57&08.66 \\
&{\checkmark}& 13.11&00.03&30.60&06.87&04.73&01.67&03.78&08.68\\\hline
\multirow{2}[0]{*}{Trans4PASS~\cite{zhang2022bending}}&&16.52&00.03&32.23&10.19&05.31&00.16&05.62&10.01 \\
&{\checkmark}&15.95&00.21&31.71&08.34&06.01&00.35&06.41&09.85\\\hline
\multirow{2}[0]{*}{UniDAPS \cite{zhang2022UniDAPS}} &&02.31&00.00&11.25&06.17&02.80&00.00&01.50& 03.43\\
&{\checkmark}&\textit{n.a.}&\textit{n.a.}&\textit{n.a.}&\textit{n.a.}&\textit{n.a.}&\textit{n.a.}&\textit{n.a.}&\textit{n.a.}\\
\hline
\multirow{2}[0]{*}{EDAPS~\cite{saha2023edaps}}&&16.61&00.04&30.83&06.49&11.42&00.37&06.19&10.28\\
&{\checkmark}&15.78&00.09&30.02&09.00&12.20&00.36&07.35&\textbf{10.68} \\\hline
\multirow{2}[0]{*}{Source-Only}&&15.79&00.02&33.88&13.54&03.58&00.19&06.78&10.54 \\
&{\checkmark}&15.18&00.01&33.37&12.60&03.28&00.35&06.77&10.22 \\\hline
\cellcolor[gray]{.9}UnmaskFormer
&\cellcolor[gray]{.9}&\cellcolor[gray]{.9}17.55&\cellcolor[gray]{.9}00.00&\cellcolor[gray]{.9}35.18&\cellcolor[gray]{.9}14.18&\cellcolor[gray]{.9}02.92&\cellcolor[gray]{.9}00.77&\cellcolor[gray]{.9}07.14&\cellcolor[gray]{.9}\textbf{11.10}\\
\cellcolor[gray]{.9}(Ours)&\cellcolor[gray]{.9}{\checkmark}&\cellcolor[gray]{.9}16.10&\cellcolor[gray]{.9}00.07&\cellcolor[gray]{.9}34.07&\cellcolor[gray]{.9}12.29&\cellcolor[gray]{.9}02.19&\cellcolor[gray]{.9}00.61&\cellcolor[gray]{.9}08.15&\cellcolor[gray]{.9}10.50\\
\bottomrule
\end{tabular}
}
\label{table:uda_sota_amodal_instance}
\end{table}
\endgroup

\cref{table:uda_sota_amodal_instance} additionally presents the results in amodal instance segmentation of our comparative experiments on the benchmark of KITTI360-APS$\rightarrow$BlendPASS transfer.
The instance segmentation performance measured in both AP for visible regions and AAP for full regions are listed.
UnmaskFormer reaches a new record of $11.10\%$ in mAP and reaches an on-par mAAP score of $10.50\%$ as EDAPS~\cite{saha2023edaps} specifically designed for UDA panoptic segmentation. 
Compared to EDAPS, our UnmaskFormer has large gains on challenging classes like \emph{trucks} and \emph{cars} frequently under large occlusions in unconstrained scenes.

\subsection{Results of Panoramic Semantic Segmentation}
We further investigate the generalization capacity of the proposed UnmaskFormer using two panoramic semantic segmentation datasets~\cite{zhang2022behind, ma2021densepass}.

\noindent\textbf{SynPASS.}
In \cref{table:synpass_benchmark}, we conduct a comparison between known convolutional and transformer models for panoramic semantic segmentation on the SynPASS~\cite{zhang2022behind}.
Here, the models are learned on the training set and evaluated on the validation set with diverse weather and illumination conditions.
UnmaskFormer equipped with the UA and the semantic head, produces SOTA performance in mIoU, which reaches $45.34\%$.
Moreover, in each adverse condition, UnmaskFormer yields robust segmentation of the full $360^{\circ}{\times}180^{\circ}$ panoramas on SynPASS. 
\begin{table}
\begin{minipage}[t]{0.6\linewidth}
\centering
\caption{\textbf{Panoramic Semantic Segmentation} results on the SynPASS benchmark~\cite{zhang2022behind}, covering four adverse weather conditions, and day- and night-time scenes.}
\hfill
\resizebox{\linewidth}{!}{
\begin{tabular}{l|cccc|cc|c} 
\toprule
\textbf{Method} & \textbf{Cloudy} & \textbf{Foggy} & \textbf{Rainy} & \textbf{Sunny} & \textbf{Day}& \textbf{Night}& {\textbf{ALL}} \\ 
\midrule\midrule
Fast-SCNN~\cite{poudel2019fastscnn} &30.84&22.68&26.16&27.19&29.68&24.75&26.31 \\
DeepLabv3+ (MNv2)~\cite{chen2018deeplabv3+} &38.94&35.19&35.43&37.73&36.01&30.55&36.72 \\
HRNet (W18Small)~\cite{wang2021hrnet} &42.92&37.94&37.37&41.45&39.19&32.22&39.80 \\ \midrule
PVT (Small)~\cite{wang2021pvt} & 40.75&36.14&34.29&40.14&37.92&28.80&37.47 \\
SegFormer (B2)~\cite{xie2021segformer} & 46.07&40.99&40.10&44.35&44.08&33.99&42.49 \\
Trans4PASS (Small)~\cite{zhang2022bending} & 46.74&43.49&43.39&45.94&45.52&37.03&44.80 \\
\rowcolor[gray]{.9}UnmaskFormer (Ours) & 48.00&43.43&43.48&47.06&45.87&36.43&\textbf{45.34} \\
\bottomrule
\end{tabular}
}
\label{table:synpass_benchmark}
\end{minipage}
\begin{minipage}[t]{0.4\linewidth}
\caption{\textbf{Results on the DensePASS benchmark}~\cite{ma2021densepass}\textbf{.} Efficiency is compared in terms of FLOPs (G) and \#Parameters (M).}
\resizebox{\linewidth}{!}{
\begin{tabular}{l|cc|c}
\toprule 
Method & FLOPs & \#Parameters & mIoU \\
\midrule\midrule
SegFormer~\cite{xie2021segformer} & 13.27 & 13.66 & 39.02 \\
DPPASS~\cite{zheng2023both_style_distortion} & \textit{n.a.} & \textit{n.a.} & 42.40 \\
DATR~\cite{zheng2023look_neighbor} & \textit{n.a.} & 14.72 & 42.22 \\
FAN~\cite{zhou2022fan} & 10.96 & 13.81 & 42.54 \\
PoolFormer~\cite{yu2022metaformer} & 09.47 & 13.47 & 43.18 \\
SegNeXt~\cite{guo2022segnext} & 14.03 & 13.71 & 43.75 \\
Trans4PASS~\cite{zhang2022bending} & 12.02 & 13.93 & 45.89 \\
\rowcolor[gray]{.9}UnmaskFormer (Ours) & 11.73 & 13.96 & \textbf{48.08} \\
\bottomrule
\end{tabular}
}
\label{tab:efficiency}
\end{minipage}

\end{table}

\noindent\textbf{DensePASS.}
As shown in \cref{tab:efficiency}, we further compare UnmaskFormer against efficient panoramic segmentation transformers including DPPASS~\cite{zheng2023both_style_distortion}, DATR~\cite{zheng2023look_neighbor}, and Trans4PASS~\cite{zhang2022behind}, where the models are trained on Cityscapes~\cite{cityscapes} and tested on the original DensePASS~\cite{ma2021densepass}.
Compared to the DATR~\cite{zheng2023look_neighbor} constituted by distortion-aware attention 
with $14.72M$ parameters, our UnmaskFormer with fewer parameters ($13.96M$) attains higher segmentation performance.
Our UnmaskFormer achieves $48.08\%$ in mIoU, which outperforms its counterparts and maintains as a lightweight segmenter.

\subsection{Ablation Study}
\label{sec:ablation_study}
To better understand the components of the UnmaskFormer, we conduct ablation studies on the KITTI360-APS$\rightarrow$BlendPASS benchmark.

\begin{figure}
    \begin{minipage}[t]{0.51\linewidth}
    \centering
    \includegraphics[width=\linewidth]{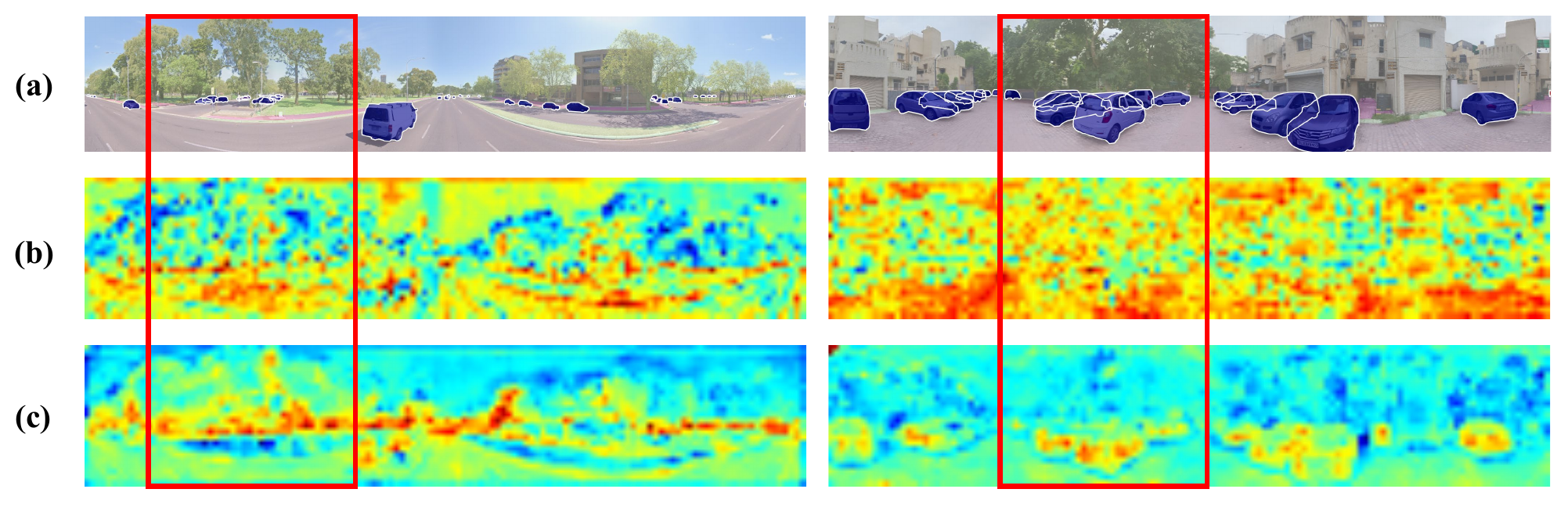}
    \caption{Features Visualization before and after UA. (a) Input image and ground truth, (b) Features before UA, and (c) Features after UA. Zoom in for a better view.}
    \label{fig_UA_vis}    
    \end{minipage}
    \hspace{2ex}
\begin{minipage}[t]{0.45\linewidth}
    \centering
    \includegraphics[width=0.9\linewidth]{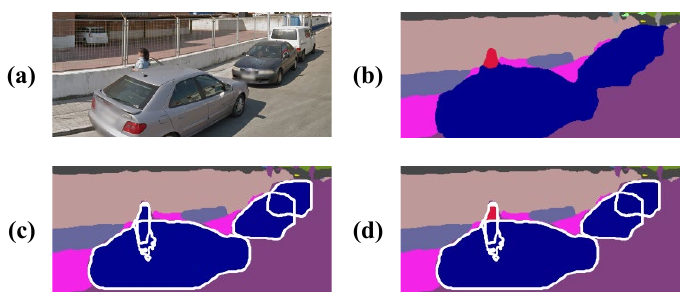}
    \caption{Visualization of different fusion strategies for APS. (a) Image, (b) Predicted semantic map, (c) Conventional fusion~\cite{saha2023edaps, cheng2020panoptic}, and (d) OAFusion.}
    \label{fig_fusion}
    \end{minipage}\hspace{3ex}
\end{figure}

\noindent \textbf{Why UA unmasks occlusions?}
In \cref{tab:ablation_UA}, we ablate different patch embedding methods~\cite{xie2021segformer, zhang2022bending} and pooling methods~\cite{psomas2023simpool}. 
The AvgPool and SimPool are used to replace the GAP operator in UA.
Compared to the original PE design and the early-stage DPE, we construct the backbone by using a new interleaving arrangement which improves the distortion-aware modeling, yielding the best mAPQ score $26.58\%$ with a ${+}3.04\%$ and a ${+}2.14\%$ gain respectively.
Compared to SimPool~\cite{psomas2023simpool}, our occlusion-aware pooling attention better handles object occlusion and obtains higher results in OASS. 
An example of feature visualization in \cref{fig_UA_vis} shows that our UA can enhance distortion-tolerant predictions of regions where occlusion occurs.
These results show the effectiveness of the UA backbone in addressing image distortion and object occlusion at the same time.

\begin{table}
\centering
\resizebox{\linewidth}{!}{
\begin{tabular}{@{}ccc@{}}
\begin{minipage}[t]{0.3\linewidth}
    
  \centering
  \caption{Unmasking Attention (UA) study.}
  \resizebox{\linewidth}{!}{
    \begin{tabular}{c|cc}
    \toprule
    Method      & \multicolumn{1}{l}{mIoU} & \multicolumn{1}{l}{mAPQ} \\
    \midrule \midrule
    PE~\cite{xie2021segformer}    & 41.07 & 22.00 \\
    DPE~\cite{zhang2022bending}   & 40.70 & 22.90 \\
    AvgPool & 42.10 & 23.90 \\
    SimPool~\cite{psomas2023simpool} & 43.04 & 24.74 \\
    \cellcolor[gray]{.9}UA    & \cellcolor[gray]{.9}43.06 & \cellcolor[gray]{.9}25.04 \\
     \bottomrule
    \end{tabular}
    }
  \label{tab:ablation_UA}
\end{minipage}
&
\begin{minipage}[t]{0.3\linewidth}
    
  \centering
  \caption{Amodal-orien-ted Mix (AoMix) study.}
  \resizebox{\linewidth}{!}{
    \begin{tabular}{c|cc}
    \toprule
    Method      & \multicolumn{1}{l}{mIoU} & \multicolumn{1}{l}{mAPQ} \\
    \midrule \midrule
    T for S & 42.53 & 23.98 \\
    T for M & 43.18 & 24.78 \\
    P for S\&M & 42.52 & 21.87 \\
    W for S\&M & 41.64 & 24.12 \\
    \cellcolor[gray]{.9}AoMix & \cellcolor[gray]{.9}42.39 & \cellcolor[gray]{.9}25.17 \\
     \bottomrule
    \end{tabular}%
    }
  \label{tab:ablation_AoMix}%
\end{minipage}
&
\begin{minipage}[t]{0.4\linewidth}
  \centering
  \caption{Ablation study of UnmaskFormer components.}
  \resizebox{\linewidth}{!}{
    \begin{tabular}{ccc|cc}
    \toprule
    \multicolumn{1}{l}{UA} & \multicolumn{1}{l}{AoMix} & \multicolumn{1}{l}{OAFusion} & \multicolumn{1}{l}{mIoU} & \multicolumn{1}{l}{mAPQ} \\
    \midrule \midrule
          &       &    \checkmark   & 41.07 & 22.00 \\
       \checkmark   &       &    \checkmark   & 43.06 & 25.04 \\
          &    \checkmark   &   \checkmark    & 42.39 & 25.17 \\
      \checkmark    &   \checkmark    &      & 43.66 & 26.55 \\
      \cellcolor[gray]{.9}\checkmark    &   \cellcolor[gray]{.9}\checkmark    &    \cellcolor[gray]{.9}\checkmark   & \cellcolor[gray]{.9}43.66 & \cellcolor[gray]{.9}26.58 \\
    \bottomrule
    \end{tabular}%
    }
  \label{tab:ablation_all}%
\end{minipage}
\end{tabular}
}
\label{tab:ablation}
\end{table}

\noindent \textbf{How AoMix boosts amodal prediction?}
As shown in \cref{tab:ablation_AoMix}, we investigate various approaches to masked image modeling. 
Compared to using masked images only for the source image (T for S) or the mixed image (T for M), our strategy AoMix, which utilizes both, achieves the best results. 
Masking the \textit{Thing} class regions (AoMix), as opposed to the entire image (W for S\&M) or masked by patches (P for S\&M), results in improvements with a gain of ${+}1.05\%$ and ${+}3.3\%$ in mAPQ respectively.
With the carefully devised AoMix, the UnmaskFormer earns significant gains by greatly boosting surrounding parsing seamlessly.

\noindent \textbf{What OAFusion fuse?}
As illustrated in \cref{fig_fusion}, a substantial portion of the pedestrian area (in \textbf{\textcolor[rgb]{0.863,0.078,0.235}{red}}) is occluded by the car (in \textbf{\textcolor[rgb]{0,0,0.557}{blue}}). 
Despite the amodal instance branch being capable of predicting the full region of the pedestrian, different fusion approaches yield distinct category classifications. 
The conventional fusion strategy~\cite{saha2023edaps, cheng2020panoptic} tailored for panoptic segmentation, given the substantial occlusion of the pedestrian region by the car, misclassifies it as \textit{car} under the majority voting rule. 
Conversely, our designed OAFusion ensures accurate classification of the full region as the \textit{pedestrian} by disregarding overlapping regions. 

Finally, as shown in \cref{tab:ablation_all}, all components cooperate with each other to achieve the best performance. More analyses can be found in the supplementary.

\subsection{Analysis of Hyperparameters}
We further conduct an analysis of relevant hyperparameters in UnmaskFormer on the KITTI360-APS$\rightarrow$BlendPASS benchmark.

\begin{figure}
    \centering
    \begin{minipage}[t]{.49\columnwidth}
    \centering
     \subfloat[mIoU]{\includegraphics[width=0.39\textwidth]{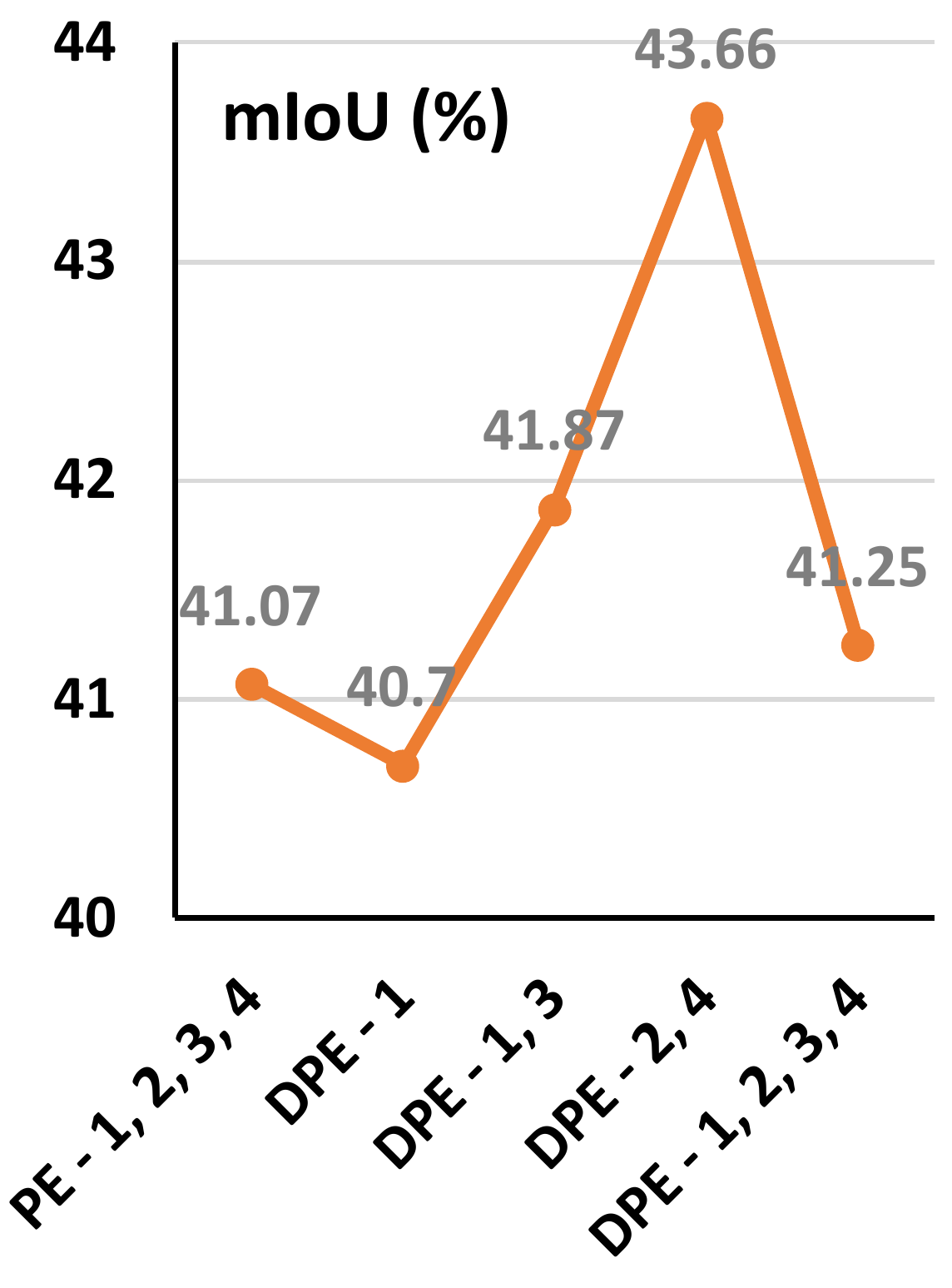}}
     \subfloat[mAPQ]{\includegraphics[width=0.39\textwidth]{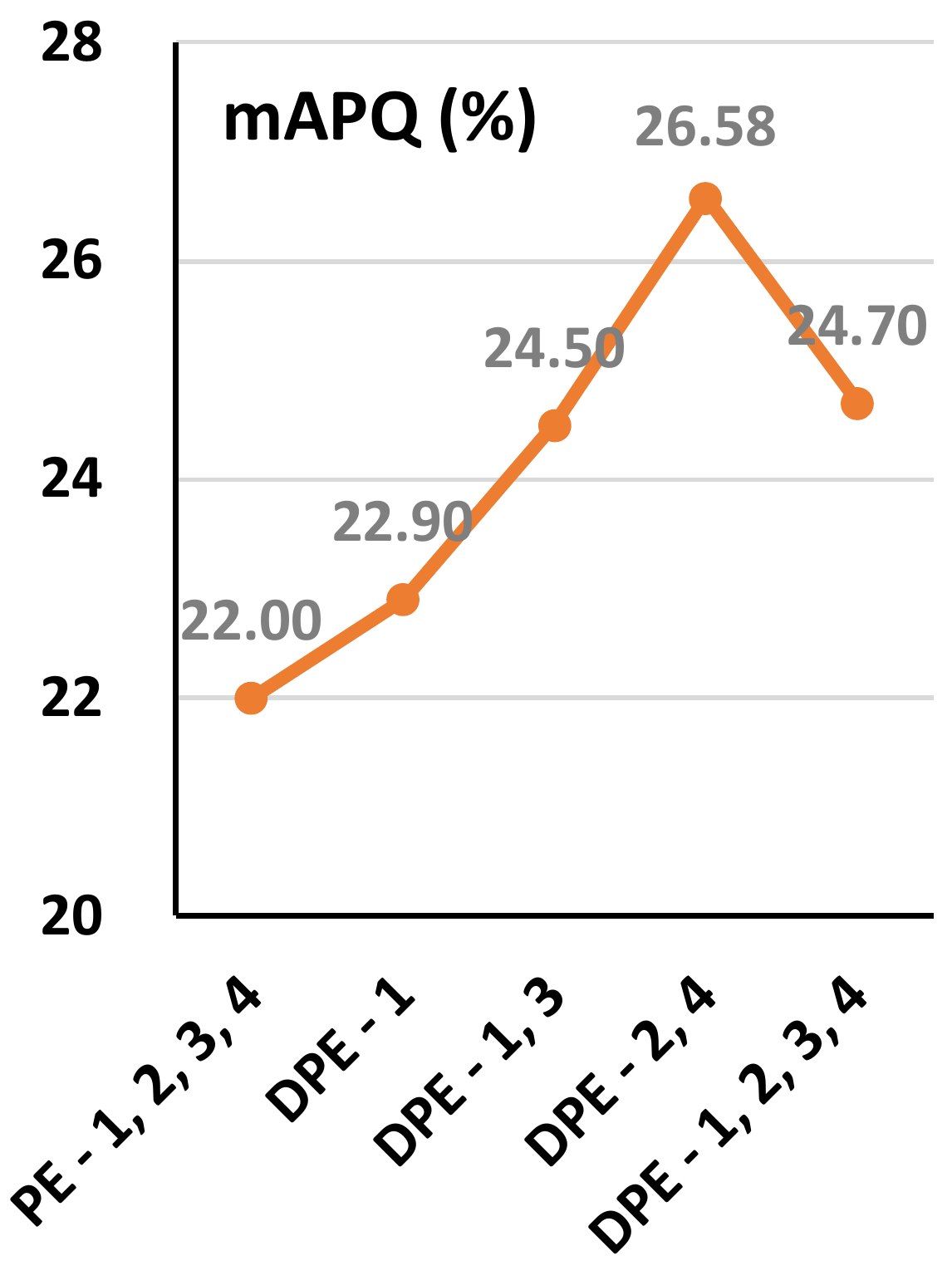}}
    \caption{Analysis of different designs with PE and DPE blocks in UnmaskFormer.}
    \label{fig:ablation_DPE}
    \end{minipage}
    {}
    \begin{minipage}[t]{.49\columnwidth}
    \centering
    \subfloat[mIoU\hfill]{\includegraphics[width=0.48\textwidth]{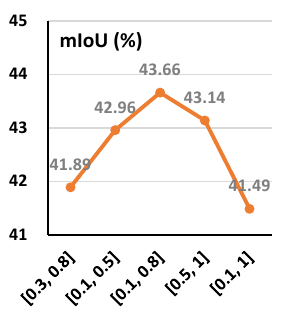}}\hfill
     \subfloat[mAPQ\hfill]{\includegraphics[width=0.48\textwidth]{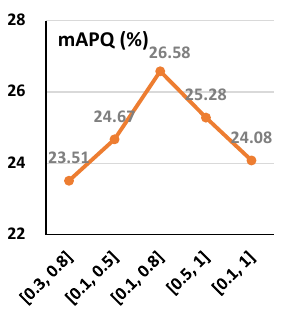}}\hfill
    \caption{Analysis of different scale ranges in the proposed AoMix method.}
    \label{fig:ablation_RS}
    \end{minipage}
\end{figure}

\noindent \textbf{Analysis of deformable designs.}
In \cref{fig:ablation_DPE}, we conduct experiments to analyze designs of  PE~\cite{xie2021segformer} and DPE~\cite{zhang2022bending} in our UnmaskFormer model.
We obtain three insights: (1) The PE cannot effectively handle the image distortion in OASS, which is in line with the finding in~\cite{zhang2022bending}. 
(2) Using more DPE blocks inside a four-stage model does not ensure optimal performance. 
(3) Compared with shallow stages, DPE can bring more improvements when acting on deep stages, 
\eg, using DPE in Stages 2 and 4 (DPE - 2, 4) performs better than in Stages 1 and 3 (DPE - 1, 3). 
Therefore, we adopt the deformable design of DPE in Stages 2 and 4 to construct the UnmaskFormer as default, which provides the optimal architecture for addressing image distortion and object deformation in OASS.

\noindent \textbf{Analysis of AoMix parameters.} 
In \cref{fig:ablation_RS}, we conduct experiments to analyze the impact of different scaling parameters in $RS(\cdot)$. 
Excessively large scaling sizes lead to complete occlusion of objects in images, preventing the model from learning original object information. 
Overly small scaling sizes result in minimal object occlusion, limiting the model's capacity to make reasonable predictions for occluded regions. Optimal performance is achieved by setting the range in $RS(\cdot)$ to $[0.1, 0.8]$, effectively generating diverse masked source images and masked mixed images. This variety of amodal-oriented samples bridges the gap between the pinhole and panoramic domains, enhancing the model's ability to reconstruct invisible regions of occluded objects and boost seamless segmentation.

\section{Conclusion}
\label{conclusion}
In this work, we have introduced the task of \textit{Occlusion-Aware Seamless Segmentation (OASS)} for holistic scene understanding.
To address OASS, we put forward \textit{UnmaskFormer} for unmasking the narrow field of view, unmasking the occlusion of perspective, and unmasking the gap of domain seamlessly.
We establish the \textit{BlendPASS} dataset for facilitating the optimization and evaluation of OASS models, as well as fostering future research in the panoramic and panoptic vision field.
Experiments on the fresh BlendPASS, as well as public SynPASS and DensePASS benchmarks, demonstrate the effectiveness of the proposed methods.

\appendix

\setcounter{table}{0}
\renewcommand{\thetable}{S\arabic{table}}

\setcounter{figure}{0}
\renewcommand{\thefigure}{S\arabic{figure}}

\section{Datasets}

\noindent\textbf{Dense annotation of BlendPASS.}
To enhance the accuracy of dense annotations for objects of the \textit{Thing} class, particularly in representing invisible regions of occluded objects realistically, our labeling process follows the ``independent annotation ${\rightarrow}$ cross-verification ${\rightarrow}$ voting'' workflow. 
All images are densely annotated by three skilled annotators using the EISeg tool for initial segmentation. Specifically, three annotators densely annotate the full region of each object, and occluded objects are independently annotated by three annotators. Subsequently, cross-verification is conducted among the annotators. In cases where there are slight discrepancies in the annotations of occluded objects, the final annotation is determined through majority voting. For annotations that do not reach a consensus, more iterative annotation processes are employed until an agreement is reached. This annotation workflow aims to ensure the accuracy of dense annotations while maximizing the fidelity of annotations for occluded regions to the true object shape.
Annotating panoramas with severe distortion is challenging and extremely time-consuming, requiring approximately $210$ minutes per person per image.
Finally, as illustrated in \cref{BlenPAS_b}, $2,960$ objects are annotated in the \textit{Thing} class. We establish a finely labeled dataset, BlendPASS, based on panoramic images containing semantic, instance, and amodal instance labels. All annotations are cross-checked to support five tasks simultaneously: semantic segmentation, instance segmentation, amodal segmentation, panoptic segmentation, and amodal panoptic segmentation.

 \begin{table}[]
     \centering
         \caption{\textbf{Statistic} for occluded and unoccluded objects in different classes.}
         \resizebox{\linewidth}{!}{
    \begin{tabular}{c|cccccccc|c}
    
        \midrule
          & {Person} & {Rider} & {Car} & {Truck} & {Bus} & {Train} & {Motorcycle} & {Bicycle} & {Total} \\
        \midrule    \midrule
    {\tabincell{c}{\#Occluded objects}} & {189} & {6} & {909} & {42} & {18} & {1} & {83} & {38} & {1286} \\
    \tabincell{c}{\#Unoccluded objects} & {613} & {12} & {842} & {38} & {24} & {2} & {71} & {72} & {1674} \\
    {Total} & {802} & {18} & {1751} & {80} & {42} & {3} & {154} & {110} & \textbf{2960} \\
        \midrule
    \end{tabular}%
    }
    \label{BlenPAS_b}

  \end{table}

\begingroup
\begin{table}[!h]
\centering
\caption{Comparison of the proposed BlendPASS with existing datasets.}
\resizebox{\linewidth}{!}{
\begin{tabular}{l|c|ccc|c}
\hline
&Panoramic Image &Semantic GT& Instance GT &Amodal Instance GT&Cross-checking\\
\hline
    \hline
KINS~\cite{qi2019amodal_kins}&\xmark&\xmark&\cmark&\cmark&\cmark \\
KITTI360-APS~\cite{mohan2022amodal_panoptic_segmentation}&\xmark&\cmark&\cmark&\cmark&\xmark \\
DensePASS~\cite{ma2021densepass}&\cmark&\cmark&\xmark&\xmark&\xmark\\
BlendPASS (Ours)&\cmark&\cmark&\cmark&\cmark&\cmark\\
\hline
\end{tabular}
}
\label{tab:Re_as}
\end{table}
\endgroup

\noindent\textbf{SynPASS.}
The SynPASS dataset~\cite{zhang2022behind} is a panoramic semantic segmentation dataset captured via the Carla simulator~\cite{dosovitskiy2017carla}.
It has four weather conditions including sunny, cloudy, foggy, and rainy scenes, together with daytime and nighttime situations.
Overall, it has $9,800$ panoramic images with a resolution of $2048{\times}1024$ corresponding to a full Field of View (FoV) of $360^\circ{\times}180^\circ$, divided into training/validation/testing sets of $5,700/1,690/1,690$ images, respectively.

\noindent\textbf{Cityscapes$\rightarrow$DensePASS}. The Cityscapes$\rightarrow$DensePASS benchmark~\cite{ma2021densepass} measures the performance of pixel-wise semantic segmentation models learned on $2,979$ labeled pinhole images of Cityscapes and transferred to DensePASS, which has $100$ panoramic images for evaluation.
The pinhole images have a resolution of $2048{\times}1024$ and the panoramic images have a resolution of $2048{\times}400$.

\noindent\textbf{KITTI360-APS$\rightarrow$BlendPASS.}
We carefully reviewed the accessible annotations for KITTI360-APS~\cite{mohan2022amodal_panoptic_segmentation}, and $12,320$ images containing $89,938$ objects of thing classes are available. 
This dataset serves as the source pinhole domain of our proposed benchmark. 
For the introduced fresh BlendPASS dataset, considering the scalability of future work, we annotated $100$ images containing $2,960$ objects of thing classes in the evaluation set, following the format of Cityscapes~\cite{cityscapes}. This serves as the validation set for the target panoramic domain of the benchmark. However, due to inconsistent classes between the two datasets, we conducted additional manual processing. Specifically, in the KITTI360-APS, there are $11$ valid \textit{Stuff} and $7$ valid \textit{Thing} classes (while the work~\cite{mohan2022amodal_panoptic_segmentation} claims $10$ classes for \textit{Stuff}, our manual verification confirmed that the \textit{traffic light} class is an additional usable annotation class).
These $11$ \textit{Stuff} classes align with BlendPASS. As for \textit{Thing} classes, we adjusted the annotations of BlendPASS to align with KITTI360-APS, following the corresponding scheme in \cref{tab:my_label}. This rough alignment further magnifies the challenges of cross-domain in the benchmark. Finally, for the OASS benchmark, there are a total of $11$ aligned \textit{Stuff} and $7$ aligned \textit{Thing} classes in both domains. 
\begin{table}[h]
    \centering
    \caption{\textbf{Alignment scheme} of the \textit{Thing} categories between the KITTI360-APS dataset and our BlendPASS dataset.}
    \resizebox{\linewidth}{!}{
    \begin{tabular}{c|cccccccc}
    \midrule
    \textbf{Dataset}&\multicolumn{8}{c}{\textbf{Categories}}\\
     \midrule
    \midrule
    BlendPASS& {Person} & {Rider} & {Car} & {Truck} & {Bus} & {Train} & {Motorcycle} & {Bicycle} \\
    KITTI360-APS~\cite{mohan2022amodal_panoptic_segmentation}& {Pedestrains} & {Cyclists} & {Car} & {Truck} & {Other-vehicles} & {Van}&{Two-wheeler} & {Two-wheeler} \\

    \midrule
    \end{tabular}%
    }

    \label{tab:my_label}
\end{table}

\section{Evaluation Metrics}
In the context of the Occlusion-Aware Seamless Segmentation (OASS) benchmark, we employ five metrics, namely Intersection over Union (IoU), Average Precision (AP), Amodal Average Precision (AAP), Panoptic Quality (PQ), and Amodal Panoptic Quality (APQ), to evaluate the model's performance. We provide a detailed explanation:

\noindent\textbf{IoU.}
IoU measures the overlap between predicted segment $p$ and ground truth segment $g$ and is calculated as: 
\begin{equation}
\text{IoU} = {(p \cap g)}/{(p\cup g)}.
\end{equation}

\noindent\textbf{AP.}
We follow Pascal VOC and COCO and use average precision, which is computed by averaging the ten equally spaced IoU thresholds from $0.5$ to $0.95$.

\noindent\textbf{AAP.}
The AAP is an extended metric of AP for Amodal Instance Segmentation, where the ground truth is replaced with amodal segments.

\noindent\textbf{PQ.}
We adapt the standard panoptic quality metric proposed by \cite{kirillov2019panoptic} and compute it as 
\begin{equation}
    \text{PQ} = \frac{\sum_{(p,g)\in {TP}}\text{IoU}(p,g)}{|TP|+ \frac{1}{2}|FP| + \frac{1}{2}|FN|},
\end{equation}
where $TP = \{(p,g)\in \boldsymbol{p}\times \boldsymbol{g}: \text{IoU}(p,g)>0.5\}$ is a set of True Positive matches.

\noindent\textbf{APQ.}
The APQ is an extended metric of PQ for Amodal Panoptic Segmentation, where the ground truth segments $\boldsymbol{g}$ are replaced with amodal segments.

\section{More Results}
\subsection{More Experiment Details}
We use the same data augmentation parameters as DACS~\cite{tranheden2021dacs} and set the RCS temperature to $0.01$ to enhance sample frequency for rare classes of the source domain. 
The EMA decay $\eta$ is set as $0.999$.
The threshold $\tau$ in pseudo-label weight is set as $0.968$.
Moreover, the pseudo-labels in the regions $11$ pixels above and $88$ pixels below the trained patches are ignored, respectively.
Moreover, we adopt the ImageNet feature distance loss from DAFormer with a weight of $0.005$. 
For instance and amodal instance branches, we set all loss weights to $1$.
During inference, the thresholds for instance and amodal instance are set to $0.95$.
For the retraining of the existing methods~\cite{saha2023edaps,zhang2022UniDAPS,zhang2022bending,zheng2023look_neighbor}, we utilize training protocols similar to ours and followed their hyperparameters to ensure fairness.
Our experiments are conducted on an NVIDIA Tesla V100 GPU and implemented using PyTorch.

\subsection{More OASS Results}
\begingroup
\begin{table*}[ht!]
\normalsize
\centering
\caption{
\textbf{Panoptic Segmentation} results on the KITTI360-APS $\rightarrow$ BlendPASS benchmark. The per-class results are reported as PQ, and the metric is mPQ.}
\resizebox{\linewidth}{!}{
\begin{tabular}{c  |l|cccccccccccccccccc|c}
\toprule 
 \multicolumn{1}{c}{Task}&\multicolumn{1}{l}{UDA Method} & \rots{road} & \rots{sidewalk} & \rots{building} & \rots{wall} & \rots{fence} & \rots{pole} & \rots{traffic-light} & \rots{traffic-sign} & \rots{vegetation} & \rots{terrain} & \rots{sky} & \rots{pedestrians} & \rots{cyclists} & \rots{car} & \rots{truck} & \rots{other-vehicles} & \rots{van} & \rots{two-wheeler} &  Metric \\
\midrule\midrule

\multirow{6}[0]{*}{\textbf{PS}}
&DATR~\cite{zheng2023look_neighbor} &50.44&09.14&59.92&11.93&11.98&01.95&00.00&03.91&64.60&14.05&70.45&12.15&00.00&38.09&00.00&03.38&00.00&01.29&19.63\\
&Trans4PASS~\cite{zhang2022bending} &53.93&14.12&69.39&19.16&11.77&03.77&00.00&05.15&67.62&16.02&77.41&14.60&04.09&38.23&06.91&00.00&00.00&07.19&22.80\\
&UniDAPS \cite{zhang2022UniDAPS} &65.95&9.48&66.31&17.39&14.28&04.77&00.00&06.14&67.19&16.10&72.68&08.27&00.00&27.25&14.82&09.23&00.00&08.86&22.71\\
&EDAPS~\cite{saha2023edaps} &55.01&17.05&66.84&18.72&14.49&05.76&04.04&04.68&68.21&16.04&72.76&19.56&00.00&37.83&01.82&04.38&00.00&07.89&23.06\\
&Source-Only&57.84&14.21&73.83&15.49&07.59&00.67&00.00&10.40&58.30&12.39&83.15&14.85&00.00&39.06&05.96&00.00&00.00&07.68&22.30\\
&\cellcolor[gray]{.9}UnmaskFormer (Ours) &\cellcolor[gray]{.9}61.73&\cellcolor[gray]{.9}24.72&\cellcolor[gray]{.9}66.80&\cellcolor[gray]{.9}20.75&\cellcolor[gray]{.9}15.81&\cellcolor[gray]{.9}05.22&\cellcolor[gray]{.9}04.29&\cellcolor[gray]{.9}03.26&\cellcolor[gray]{.9}69.02&\cellcolor[gray]{.9}18.35&\cellcolor[gray]{.9}79.44&\cellcolor[gray]{.9}20.90&\cellcolor[gray]{.9}03.45&\cellcolor[gray]{.9}42.96&\cellcolor[gray]{.9}11.26&\cellcolor[gray]{.9}07.64&\cellcolor[gray]{.9}00.00&\cellcolor[gray]{.9}15.98&\cellcolor[gray]{.9}\textbf{26.20}\\
\bottomrule
\end{tabular}
}
\label{table:uda_sota_k2d_PQ}
\end{table*}
\endgroup

In this work, we include a comprehensive set of sub-tasks, including semantic segmentation, instance segmentation, amodal instance segmentation, panoptic segmentation, and amodal panoptic segmentation for OASS. To provide a thorough analysis of these diverse tasks, we present the per-class accuracy results. \cref{table:uda_sota_k2d_PQ} further details the per-class accuracy specifically in panoptic segmentation (PS). To benchmark our model, we conduct a comparative study against previous state-of-the-art methods, namely DATR~\cite{zheng2023look_neighbor}, Trans4PASS~\cite{zhang2022bending}, UniDAPS~\cite{zhang2022UniDAPS}, and EDAPS~\cite{saha2023edaps}. As shown in the experimental results, our model demonstrates a notable Mean-PQ of $26.20\%$, surpassing the performance of the previous best model by a significant margin of ${+}3.14\%$. A detailed breakdown of per-class performance reveals that our model excels across numerous categories, such as \textit{sidewalk}, \textit{wall}, \textit{pedestrians}, \textit{car}, and \textit{two-wheeler}. It is worth noting that the \textit{van} category, constituting a minor proportion in both source and target datasets, poses a challenge for all models to handle effectively. The superior mPQ achieved by our model proves its effectiveness in addressing the challenges in the panoptic segmentation part of the OASS task.

\subsection{More Analysis of Ablation Study}
\noindent \textbf{Preliminary of Deformable Patch Embedding.}
The backbone utilizes a transformer-based structure with a novel arrangement of Deformable Patch Embedding (DPE)~\cite{zhang2022bending} layers. DPE aims to capture local geometric variations caused by image distortion. Given an input image or features $\boldsymbol{X}{\in}\mathbb{R}^{H{\times}W{\times}C}$  (where $H$ and $W$ represent the resolution and $C$ the number of channels), DPE calculates adaptive offsets for each patch:

\begin{equation}
\label{eq:dpe}
\begin{aligned}
\boldsymbol{\Delta}^{DPE}{(i,j)} &= \begin{bmatrix}
\min(\max(-\frac{H}{r}, g(\boldsymbol{X})_{(i,j)}), \frac{H}{r}) \
\min(\max(-\frac{W}{r}, g(\boldsymbol{X})_{(i,j)}), \frac{W}{r})
\end{bmatrix},
\end{aligned}
\end{equation}

\noindent where $(i,j)$ denotes the patch index, $g(\cdot)$ is the offset prediction function, $r$ is a hyperparameter controlling the maximum allowed offset.

\noindent \textbf{Analysis of UA-based Backbone.} To improve the modeling capacity of our OASS model, we take not only the deformable patchifying but also the unmasking into account. Specifically, the technical designs involve multiple perspectives. \textit{1) Interleaved DPE arrangement:} Unlike prior work~\cite{zhang2022bending} that inserts DPE only in the initial stage, this approach proposes a novel interleaving arrangement. DPE layers are strategically placed within Stages 2 and 4 of the backbone architecture. This design reinforces the model's ability to capture distortion throughout the processing pipeline. \textit{2) Unmasking Attention (UA) for occlusion handling:} In addition to addressing distortion, the method incorporates an Unmasking Attention (UA) block that is enhanced by adding a simple yet effective pooling layer. 
\textit{3) Combining self-attention and enhanced pooling:} The UA block leverages both a self-attention layer and an improved pooling mechanism. Having the self-attended pooling feature  {$\boldsymbol{q'}$}${\in}\mathbb{R}^{1{\times}1{\times}C}$, a \texttt{sigmoid} function $\phi(\cdot)$ is applied to calculate the occlusion-aware mask $\phi({{\boldsymbol{q'}}})$. This mask highlights regions likely affected by occlusion. \textit{4) Incorporating occlusion awareness:} The mask $\phi({{\boldsymbol{q'}}})$ is subsequently used to perform element-wise multiplication with the original feature map, resulting in an occlusion-aware feature. The occlusion-aware feature is further processed by an MLP layer. Based on these crucial designs, our UA-based backbone can address both the image distortion and object occlusion.  

\begin{figure}[!h]
    \centering
    \includegraphics[width=.95\linewidth]{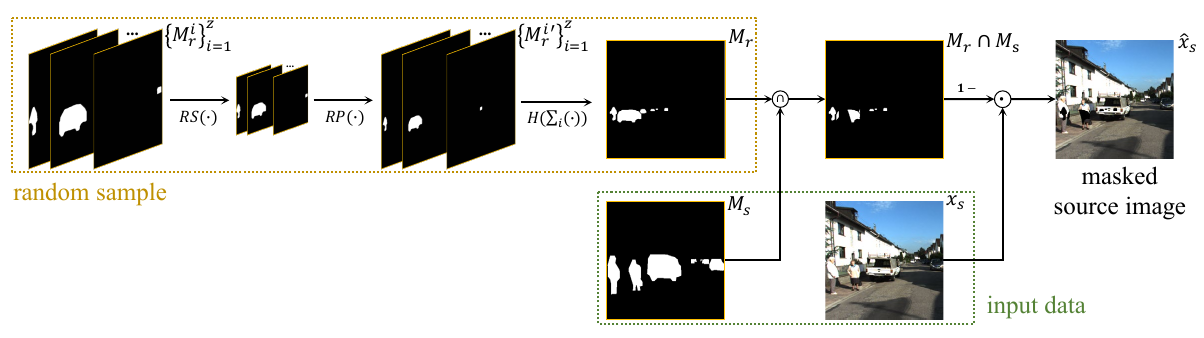}
    \caption{The process of modeling amodal-oriented masked source images.}
    \label{fig_AoMix_vis}
\end{figure}
\noindent \textbf{Analysis of AoMix.}
To provide additional amodal-oriented source priors and mixed image samples to enhance the adaptation, we utilize amodal instance masks to mask input images within the AoMix module.
An example of the amodal-oriented masked image modeling process for the source image is illustrated in \cref{fig_AoMix_vis}.
This method aims to enhance the model’s ability to reconstruct object regions obscured by realistic object shapes, enabling it to learn information about invisible parts in the scene. 
As shown in Tab. 6, we conducted an ablation study on the manner and strategy of AoMix. \textit{1) T for S, T for M vs AoMix:} Using the amodal-oriented masked image in both the source and mixed images enables the model to accurately segment the full regions of occluded objects in the source domain and facilitates better model adaptation to the target panoramic domain. \textit{2) P for S$\&$M vs AoMix:} It is noteworthy that we tried to use random patches instead of amodal instance masks to mask images. Although the mIoU score remained largely unchanged, the mAPQ score witnessed a significant drop of $3.3\%$, affirming that masks with real shapes provide better guidance for learning occlusion-ignored segmentation ability. \textit{3) W for S$\&$M vs AoMix:} Compared with applying amodal masks to all regions of the image, our method only applies to the \textit{Thing} class region, resulting in superior performance. This is because our method aligns more closely with the real-world scenario where object occlusion of the \textit{Thing} class is often raised by other objects of the \textit{Thing} class.

\section{More Visualization Results}

\subsection{More Visualization Results of OASS}

\begin{figure}
    \centering
    {
    \newcolumntype{P}[1]{>{\centering\arraybackslash}p{#1}}
    \renewcommand{\arraystretch}{1}
    \resizebox{0.96\linewidth}{!}{
    \begin{tabular}{@{}*{12}{P{0.115\columnwidth}}@{}}
    \textit{Stuff:}
    &{\cellcolor[rgb]{0.5,0.25,0.5}}\textcolor{white}{road}
    &{\cellcolor[rgb]{0.957,0.137,0.91}}sidew.
    &{\cellcolor[rgb]{0.275,0.275,0.275}}\textcolor{white}{build.}
    &{\cellcolor[rgb]{0.4,0.4,0.612}}\textcolor{white}{wall}
    &{\cellcolor[rgb]{0.745,0.6,0.6}}fence
    &{\cellcolor[rgb]{0.6,0.6,0.6}}pole
    &{\cellcolor[rgb]{0.98,0.667,0.118}}tr.light
    &{\cellcolor[rgb]{0.863,0.863,0}}tr.sign
    &{\cellcolor[rgb]{0.42,0.557,0.137}}veget.
    &{\cellcolor[rgb]{0.596,0.984,0.596}}terrain
    &{\cellcolor[rgb]{0.275,0.51,0.706}}sky\\
    
    \textit{Thing:}
    &{{\cellcolor[rgb]{0.863,0.078,0.235}}\textcolor{white}{pedes.}}
    &{\cellcolor[rgb]{1,0,0}}\textcolor{black}{cyclists}
    &{\cellcolor[rgb]{0,0,0.557}}\textcolor{white}{car}
    &{\cellcolor[rgb]{0,0,0.275}}\textcolor{white}{truck}
    &{{\cellcolor[rgb]{0,0.235,0.392}}\textcolor{white}{ot.veh.}}
    &{\cellcolor[rgb]{0,0.314,0.392}}\textcolor{white}{van}
    &{{\cellcolor[rgb]{0,0,0.902}}\textcolor{white}{tw.whe.}}\\
    \end{tabular}
    }
    }
    \includegraphics[width=\linewidth]{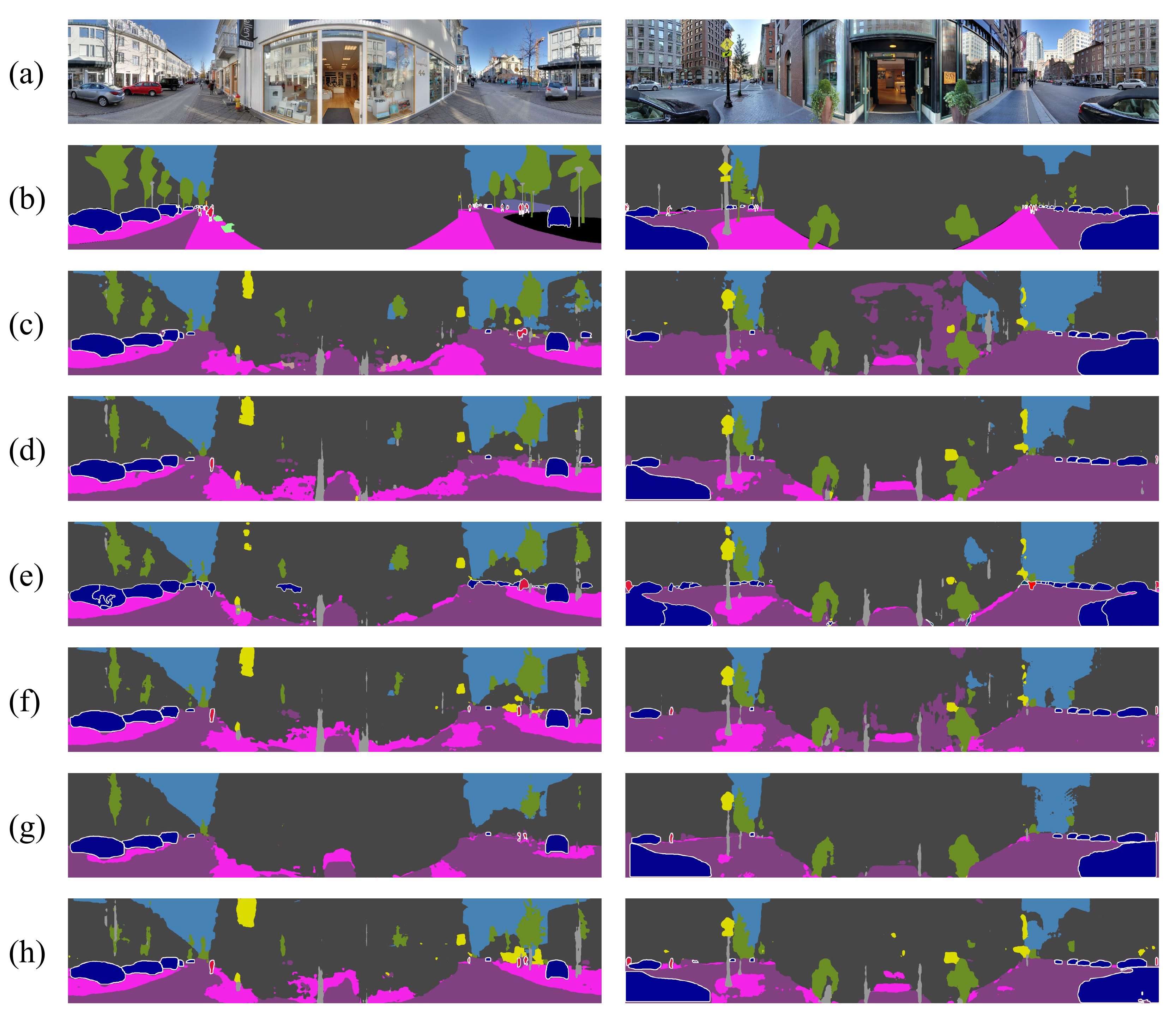}
    \caption{\textbf{Visualization results of Panoptic Segmentation.} 
    From top to bottom are
    (a) Image, (b) Ground truth, (c) DATR~\cite{zheng2023look_neighbor}, (d) Trans4PASS~\cite{zhang2022bending}, (e) UniDAPS~\cite{zhang2022UniDAPS}, (f) EDAPS~\cite{saha2023edaps}, (g) Source-Only, and (h) UnmaskFormer (Ours).}
    \label{fig_vis_panoptic_segmentation}
\end{figure}

\begin{figure}
    \centering
    
        {
    \newcolumntype{P}[1]{>{\centering\arraybackslash}p{#1}}
    \renewcommand{\arraystretch}{1}
    \resizebox{0.96\linewidth}{!}{
    \begin{tabular}{@{}*{12}{P{0.115\columnwidth}}@{}}
    \textit{Stuff:}
    &{\cellcolor[rgb]{0.5,0.25,0.5}}\textcolor{white}{road}
    &{\cellcolor[rgb]{0.957,0.137,0.91}}sidew.
    &{\cellcolor[rgb]{0.275,0.275,0.275}}\textcolor{white}{build.}
    &{\cellcolor[rgb]{0.4,0.4,0.612}}\textcolor{white}{wall}
    &{\cellcolor[rgb]{0.745,0.6,0.6}}fence
    &{\cellcolor[rgb]{0.6,0.6,0.6}}pole
    &{\cellcolor[rgb]{0.98,0.667,0.118}}tr.light
    &{\cellcolor[rgb]{0.863,0.863,0}}tr.sign
    &{\cellcolor[rgb]{0.42,0.557,0.137}}veget.
    &{\cellcolor[rgb]{0.596,0.984,0.596}}terrain
    &{\cellcolor[rgb]{0.275,0.51,0.706}}sky\\
    
    \textit{Thing:}
    &{{\cellcolor[rgb]{0.863,0.078,0.235}}\textcolor{white}{pedes.}}
    &{\cellcolor[rgb]{1,0,0}}\textcolor{black}{cyclists}
    &{\cellcolor[rgb]{0,0,0.557}}\textcolor{white}{car}
    &{\cellcolor[rgb]{0,0,0.275}}\textcolor{white}{truck}
    &{{\cellcolor[rgb]{0,0.235,0.392}}\textcolor{white}{ot.veh.}}
    &{\cellcolor[rgb]{0,0.314,0.392}}\textcolor{white}{van}
    &{{\cellcolor[rgb]{0,0,0.902}}\textcolor{white}{tw.whe.}}\\
    \end{tabular}
    }
    }
    \includegraphics[width=\linewidth]{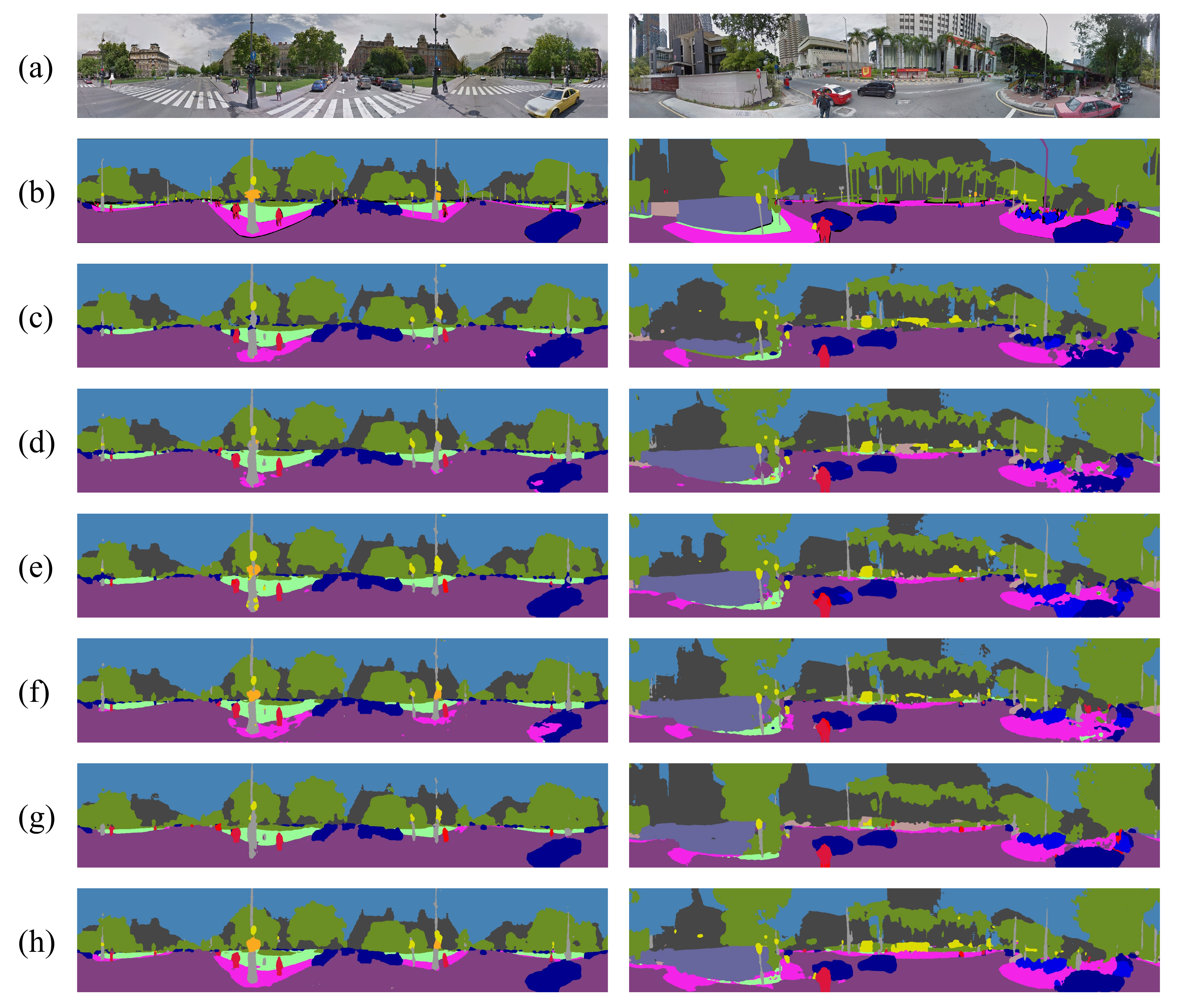}
    \caption{\textbf{Visualization results of Semantic Segmentation.} 
    From top to bottom are
    (a) Image, (b) Ground truth, (c) DATR~\cite{zheng2023look_neighbor}, (d) Trans4PASS~\cite{zhang2022bending}, (e) UniDAPS~\cite{zhang2022UniDAPS}, (f) EDAPS~\cite{saha2023edaps}, (g) Source-Only, and (h) UnmaskFormer (Ours).}
    \label{fig_vis_semantic_segmentation}
\end{figure}

We showcase visualization results for semantic segmentation and panoptic segmentation in \cref{fig_vis_panoptic_segmentation} and \cref{fig_vis_semantic_segmentation}. 
These examples show that UnmaskFormer achieves outstanding performance on other tasks within the OASS benchmark. 
In addition to addressing the occlusion of perspective, UnmaskFormer can unmask the narrow field of view and the gap of domain. 
As depicted in \cref{fig_vis_panoptic_segmentation} for panoptic segmentation, UnmaskFormer surpasses other methods~\cite{zheng2023look_neighbor,zhang2022bending,zhang2022UniDAPS,saha2023edaps} and successfully detects more pedestrians. 
Compared to the contour-based UniDAPS~\cite{zhang2022UniDAPS}, UnmaskFormer segments the \textit{Thing} objects with more complete and rational shapes. 
For the category determination of instance masks, existing methods typically rely on fusing results from the semantic branch.
Moreover, the visualization results for semantic segmentation in \cref{fig_vis_semantic_segmentation} show that UnmaskFormer achieves more accurate semantic classification results and demonstrates robustness to distortions introduced by wide-FoV panoramic images. 


\subsection{More Visualization Results on SynPASS}
\begin{figure}
    \centering
    \scriptsize
    {
    \newcolumntype{P}[1]{>{\centering\arraybackslash}p{#1}}
    \resizebox{\linewidth}{!}{
    \begin{tabular}{@{}*{11}{P{0.14\columnwidth}}@{}}
    {\cellcolor[rgb]{0.27,0.27,0.27}}\textcolor{white}{Building} 
    &{\cellcolor[rgb]{0.16,0.16,0.39}}\textcolor{white}{Fence}
    &{\cellcolor[rgb]{0.31,0.35,0.21}}\textcolor{white}{Other} 
    &{\cellcolor[rgb]{0.23,0.08,0.86}}\textcolor{white}{Pedestrian} 
    &{\cellcolor[rgb]{0.60,0.60,0.60}}\textcolor{white}{Pole}
    &{\cellcolor[rgb]{0.20,0.91,0.61}}Road~Line
    &{\cellcolor[rgb]{0.50,0.25,0.50}}\textcolor{white}{Road}
    &{\cellcolor[rgb]{0.91,0.14,0.95}}\textcolor{white}{Sidewalk}
    &{\cellcolor[rgb]{0.14,0.55,0.42}}\textcolor{white}{Vegetation}
    &{\cellcolor[rgb]{0.55,0.00,0.00}}\textcolor{white}{Vehicles}
    &{\cellcolor[rgb]{0.61,0.40,0.40}}\textcolor{white}{Wall}\\
    {\cellcolor[rgb]{0.00,0.86,0.86}}Tr. Sign
    &{\cellcolor[rgb]{0.70,0.51,0.27}}\textcolor{white}{Sky}
    &{\cellcolor[rgb]{0.32,0.00,0.32}}\textcolor{white}{Ground} 
    &{\cellcolor[rgb]{0.39,0.39,0.59}}\textcolor{white}{Bridge} 
    &{\cellcolor[rgb]{0.55,0.59,0.90}}\textcolor{white}{Rail~Track} 
    &{\cellcolor[rgb]{0.70,0.64,0.70}}\textcolor{white}{Ground~Rail} 
    &{\cellcolor[rgb]{0.12,0.66,0.98}}\textcolor{white}{Tr. Light}
    &{\cellcolor[rgb]{0.63,0.74,0.43}}\textcolor{black}{Static} 
    &{\cellcolor[rgb]{0.20,0.47,0.66}}\textcolor{white}{Dynamic} 
    &{\cellcolor[rgb]{0.59,0.23,0.18}}\textcolor{white}{Water}
    &{\cellcolor[rgb]{0.39,0.66,0.57}}\textcolor{white}{Terrain}\\
    \end{tabular}

    }
    } 
    \includegraphics[width=\linewidth]{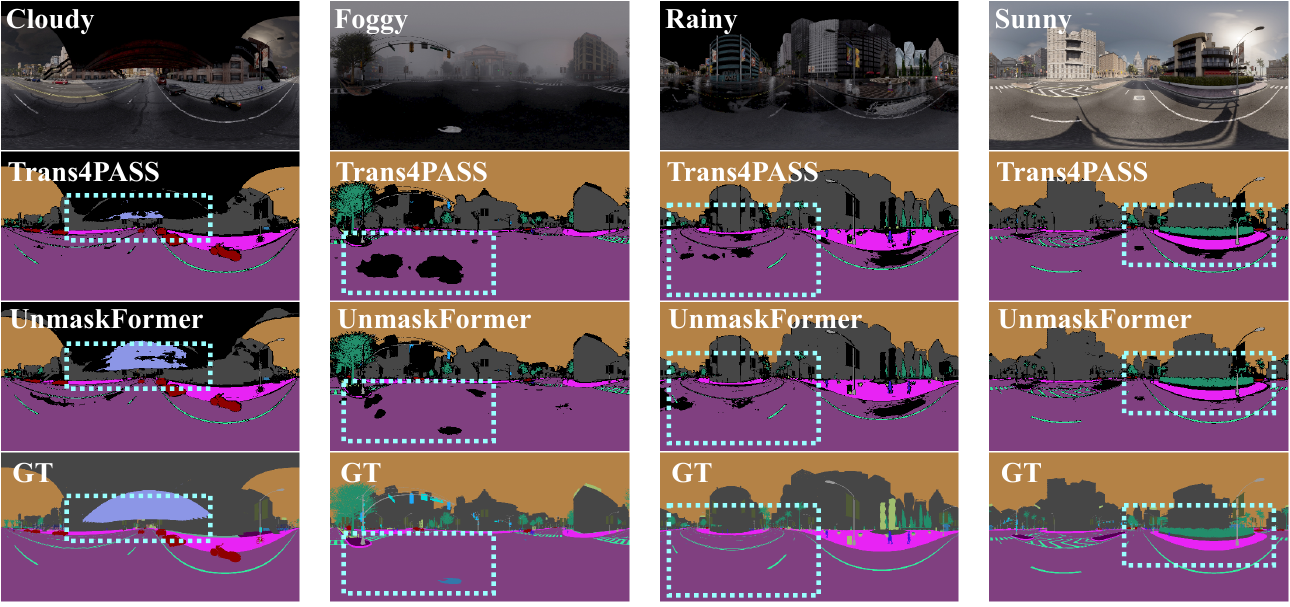}
    \caption{\textbf{More visualization results of SynPASS.} 
    From top to bottom are: Image, the prediction of Trans4PASS~\cite{zhang2022bending}, the prediction of our UnmaskFormer, and Ground truth.}
    \label{fig_vis_synpass}
\end{figure}

As shown in \cref{fig_vis_synpass}, we conduct qualitative analyses of panoramic semantic segmentation on the SynPASS dataset. 
The objective of this analysis is to evaluate the robustness of various methods under diverse weather conditions. 
Specifically, we examine four distinct weather types, namely \textit{cloudy}, \textit{foggy}, \textit{rainy}, and \textit{sunny}, to assess the adaptability of the models. 
From top to bottom in \cref{fig_vis_synpass}, they are the input image, the segmentation results of Trans4PASS~\cite{zhang2022bending}, the segmentation results of UnmaskFormer, and ground truth. 
Comparing the performance of UnmaskFormer with the prior state-of-the-art model, Trans4PASS~\cite{zhang2022bending}, it can be seen that UnmaskFormer obtains notable improvements in handling adverse weather conditions. 
For instance, in \textit{cloudy} weather, UnmaskFormer demonstrates an enhanced capability to predict and generate more complete \textit{rail track} categories. Moreover, in \textit{foggy} and \textit{rainy} weather, UnmaskFormer can identify \textit{roads} more completely compared to the baseline model. 
Thanks to the deformable designs, UnmaskFormer can yield better panoramic semantic segmentation results. 
These observations demonstrate the efficacy of UnmaskFormer in challenging environmental scenarios. 

\section{Discussion}
\textbf{Future work.}  
The advancements made in Occlusion-Aware Seamless Segmentation (OASS) through our UnmaskFormer open new avenues for future research. In the future, we envision several potential further directions:
\textit{1) User-Interactive Image Editing:} We propose exploring user interaction in image editing. Future iterations can integrate user-selected areas to separate amodal layers and employ diffusion processes for completion. These interactive features enhance user-friendliness and expand practical applications.
\textit{2) Amodal Optical Flow Fusion:} We intend to further investigate combining our segmentation approach with amodal optical flow techniques to improve temporal consistency in amodal segmentation. This is useful for dealing with temporal occlusions. Also, it is essential for continuous tracking or monitoring applications, potentially revolutionizing dynamic scene processing.

\noindent \textbf{Limitations and potential solutions.} 
While our UnmaskFormer for OASS showcases significant advancements, we acknowledge certain limitations in our current work, alongside potential approaches to address these challenges:
\textit{1) Complex Environment Perception:} 
The current model may not handle the complexity of in-the-wild scenes. 
Continuous improvement is vital for panoramic image processing and scene understanding.
We could enhance robustness through diverse data, integration with other sensors like LiDAR or event cameras, and depth-aware transferring learning to enhance occlusion reasoning. Additionally, domain generalization techniques can be employed to improve the model's adaptability to new and unseen environments, further bolstering its performance in complex scenarios.
\textit{2) Amodal Data Annotation Challenge:} 
Amodal data annotation for panoramic images is challenging.
While our dataset covers diverse images captured in cities located on all continents, 
we could further explore automated amodal annotation for panoramas, reserving manual annotation for data purification and cleaning purposes. 
Moreover, semi-supervised amodal instance segmentation methods~\cite{liu2024blade} can also be leveraged to enhance the efficiency of the annotation process.
These approaches can help scale up the availability of annotated data for training and validation, addressing the scarcity of labeled data in the amodal segmentation domain.

\noindent\textbf{Societal impacts.} 
In this study, we have introduced a novel task called Occlusion-Aware Seamless Segmentation (OASS) and established a comprehensive benchmark incorporating various well-known baseline models. 
We found that these baseline models exhibit limited performance in the OASS task, primarily due to the intricate nature of occlusion-aware segmentation challenges.
To address this, we have developed UnmaskFormer, a solution that significantly enhances performance on the OASS benchmark, outperforming existing domain adaptation panoramic and panoptic segmentation methods and achieving promising state-of-the-art results. 
Nevertheless, given the criticality of dependability in deep learning systems for Advanced Driver-Assistance Systems (ADAS), it is important to note that UnmaskFormer may still encounter misclassifications in challenging occluded regions and biased content, potentially leading to erroneous predictions with adverse societal implications.

\section*{Acknowledgements}
This work was supported in part by the Major Research Plan of the National Natural Science Foundation of China under Grant 92148204, the National Key RD Program under Grant 2022YFB4701400, the Hunan Leading Talent of Technological Innovation under Grant 2022RC3063, the Top Ten Technical Research Projects of Hunan Province under Grant 2024GK1010, the Key Research Development Program of Hunan Province under Grant 2022GK2011, and in part by Hangzhou SurImage Technology Company Ltd.
%
%
\bibliographystyle{splncs04}
\bibliography{main}

\begin{thebibliography}{10}
\providecommand{\url}[1]{\texttt{#1}}
\providecommand{\urlprefix}{URL }
\providecommand{\doi}[1]{https://doi.org/#1}

\bibitem{ai2022deep_omnidirectional}
Ai, H., Cao, Z., Zhu, J., Bai, H., Chen, Y., Wang, L.: Deep learning for omnidirectional vision: A survey and new perspectives. arXiv preprint arXiv:2205.10468  (2022)

\bibitem{ao2023image_survey}
Ao, J., Ke, Q., Ehinger, K.A.: Image amodal completion: A survey. Computer Vision and Image Understanding  (2023)

\bibitem{back2022unseen_occlusion}
Back, S., Lee, J., Kim, T., Noh, S., Kang, R., Bak, S., Lee, K.: Unseen object amodal instance segmentation via hierarchical occlusion modeling. In: ICRA (2022)

\bibitem{breitenstein2022amodal_cityscapes}
Breitenstein, J., Fingscheidt, T.: Amodal cityscapes: {A} new dataset, its generation, and an amodal semantic segmentation challenge baseline. In: IV (2022)

\bibitem{chen2023amodal_expansion}
Chen, J., Niu, L., Zhang, J., Si, J., Qian, C., Zhang, L.: Amodal instance segmentation via prior-guided expansion. In: AAAI (2023)

\bibitem{chen2018deeplabv3+}
Chen, L.C., Zhu, Y., Papandreou, G., Schroff, F., Adam, H.: Encoder-decoder with atrous separable convolution for semantic image segmentation. In: ECCV (2018)

\bibitem{chen2020banet}
Chen, Y., Lin, G., Li, S., Bourahla, O., Wu, Y., Wang, F., Feng, J., Xu, M., Li, X.: {BANet:} {Bidirectional} aggregation network with occlusion handling for panoptic segmentation. In: CVPR (2020)

\bibitem{cheng2020panoptic}
Cheng, B., Collins, M.D., Zhu, Y., Liu, T., Huang, T.S., Adam, H., Chen, L.: {Panoptic-DeepLab:} {A} simple, strong, and fast baseline for bottom-up panoptic segmentation. In: CVPR (2020)

\bibitem{cityscapes}
Cordts, M., Omran, M., Ramos, S., Rehfeld, T., Enzweiler, M., Benenson, R., Franke, U., Roth, S., Schiele, B.: The cityscapes dataset for semantic urban scene understanding. In: CVPR (2016)

\bibitem{dai2015convolutional}
Dai, J., He, K., Sun, J.: Convolutional feature masking for joint object and stuff segmentation. In: CVPR (2015)

\bibitem{deng2017cnn}
Deng, L., Yang, M., Qian, Y., Wang, C., Wang, B.: {CNN} based semantic segmentation for urban traffic scenes using fisheye camera. In: IV (2017)

\bibitem{dosovitskiy2017carla}
Dosovitskiy, A., Ros, G., Codevilla, F., Lopez, A., Koltun, V.: {CARLA:} {An} open urban driving simulator. In: CoRL (2017)

\bibitem{fan2023rethinking}
Fan, K., Lei, J., Qian, X., Yu, M., Xiao, T., He, T., Zhang, Z., Fu, Y.: Rethinking amodal video segmentation from learning supervised signals with object-centric representation. In: ICCV (2023)

\bibitem{follmann2019learning_see_invisible}
Follmann, P., K{\"o}nig, R., H{\"a}rtinger, P., Klostermann, M., B{\"o}ttger, T.: Learning to see the invisible: End-to-end trainable amodal instance segmentation. In: WACV (2019)

\bibitem{fu2023panopticnerf}
Fu, X., Zhang, S., Chen, T., Lu, Y., Zhou, X., Geiger, A., Liao, Y.: {PanopticNeRF-360:} {Panoramic 3D-to-2D} label transfer in urban scenes. arXiv preprint arXiv:2309.10815  (2023)

\bibitem{gao2023coarse}
Gao, J., Qian, X., Wang, Y., Xiao, T., He, T., Zhang, Z., Fu, Y.: Coarse-to-fine amodal segmentation with shape prior. In: ICCV (2023)

\bibitem{gao2022review}
Gao, S., Yang, K., Shi, H., Wang, K., Bai, J.: Review on panoramic imaging and its applications in scene understanding. IEEE Transactions on Instrumentation and Measurement  (2022)

\bibitem{geiger2013vision}
Geiger, A., Lenz, P., Stiller, C., Urtasun, R.: Vision meets robotics: {The KITTI} dataset. The International Journal of Robotics Research  (2013)

\bibitem{guo2022segnext}
Guo, M.H., Lu, C.Z., Hou, Q., Liu, Z., Cheng, M.M., Hu, S.M.: {SegNeXt:} {Rethinking} convolutional attention design for semantic segmentation. In: NeurIPS (2022)

\bibitem{guttikonda2023single_spherical}
Guttikonda, S., Rambach, J.: Single frame semantic segmentation using multi-modal spherical images. In: WACV (2024)

\bibitem{hariharan2014simultaneous}
Hariharan, B., Arbel{\'a}ez, P., Girshick, R., Malik, J.: Simultaneous detection and segmentation. In: ECCV (2014)

\bibitem{he2017mask}
He, K., Gkioxari, G., Doll{\'a}r, P., Girshick, R.: {Mask R-CNN}. In: ICCV (2017)

\bibitem{hoyer2022daformer}
Hoyer, L., Dai, D., Van~Gool, L.: {DAFormer:} {Improving} network architectures and training strategies for domain-adaptive semantic segmentation. In: CVPR (2022)

\bibitem{hoyer2022hrda}
Hoyer, L., Dai, D., Van~Gool, L.: {HRDA:} {Context-aware} high-resolution domain-adaptive semantic segmentation. In: ECCV (2022)

\bibitem{hoyer2023mic}
Hoyer, L., Dai, D., Wang, H., Van~Gool, L.: {MIC:} {Masked} image consistency for context-enhanced domain adaptation. In: CVPR (2023)

\bibitem{hu2022distortion}
Hu, X., An, Y., Shao, C., Hu, H.: Distortion convolution module for semantic segmentation of panoramic images based on the image-forming principle. IEEE Transactions on Instrumentation and Measurement  (2022)

\bibitem{hu2019sail}
Hu, Y.T., Chen, H.S., Hui, K., Huang, J.B., Schwing, A.G.: {SAIL-VOS:} {Semantic} amodal instance level video object segmentation - {A} synthetic dataset and baselines. In: CVPR (2019)

\bibitem{jang2022dada_uda}
Jang, S., Na, J., Oh, D.: {DaDA:} {Distortion-aware} domain adaptation for unsupervised semantic segmentation. In: NeurIPS (2022)

\bibitem{jaus2021panoramic_towards}
Jaus, A., Yang, K., Stiefelhagen, R.: Panoramic panoptic segmentation: Towards complete surrounding understanding via unsupervised contrastive learning. In: IV (2021)

\bibitem{jaus2023panoramic_insights}
Jaus, A., Yang, K., Stiefelhagen, R.: Panoramic panoptic segmentation: Insights into surrounding parsing for mobile agents via unsupervised contrastive learning. IEEE Transactions on Intelligent Transportation Systems  (2023)

\bibitem{jiang2023minimalist}
Jiang, Q., Gao, S., Gao, Y., Yang, K., Yi, Z., Shi, H., Sun, L., Wang, K.: Minimalist and high-quality panoramic imaging with {PSF-aware} transformers. IEEE Transactions on Image Processing  (2024)

\bibitem{jiang2022annular}
Jiang, Q., Shi, H., Sun, L., Gao, S., Yang, K., Wang, K.: Annular computational imaging: Capture clear panoramic images through simple lens. IEEE Transactions on Computational Imaging  (2022)

\bibitem{ke2021deep_occlusion_aware}
Ke, L., Tai, Y.W., Tang, C.K.: Deep occlusion-aware instance segmentation with overlapping bilayers. In: CVPR (2021)

\bibitem{kim2022pasts}
Kim, J., Jeong, S., Sohn, K.: {PASTS:} {Toward} effective distilling transformer for panoramic semantic segmentation. In: ICIP (2022)

\bibitem{kirillov2019panoptic}
Kirillov, A., He, K., Girshick, R., Rother, C., Doll{\'a}r, P.: Panoptic segmentation. In: CVPR (2019)

\bibitem{kirillov2023segment}
Kirillov, A., Mintun, E., Ravi, N., Mao, H., Rolland, C., Gustafson, L., Xiao, T., Whitehead, S., Berg, A.C., Lo, W., Doll{\'{a}}r, P., Girshick, R.B.: Segment anything. In: ICCV (2023)

\bibitem{lazarow2020learning_instance_occlusion}
Lazarow, J., Lee, K., Shi, K., Tu, Z.: Learning instance occlusion for panoptic segmentation. In: CVPR (2020)

\bibitem{li2016amodal_instance_segmentation}
Li, K., Malik, J.: Amodal instance segmentation. In: ECCV (2016)

\bibitem{li2023sgat4pass}
Li, X., Wu, T., Qi, Z., Wang, G., Shan, Y., Li, X.: {SGAT4PASS:} {Spherical} geometry-aware transformer for panoramic semantic segmentation. In: IJCAI (2023)

\bibitem{li20222d_3d_prior}
Li, Z., Ye, W., Jiang, T., Huang, T.: {2D} amodal instance segmentation guided by {3D} shape prior. In: ECCV (2022)

\bibitem{li2023gin}
Li, Z., Ye, W., Jiang, T., Huang, T.: {GIN:} {Generative} invariant shape prior for amodal instance segmentation. IEEE Transactions on Multimedia  (2023)

\bibitem{li2023muva}
Li, Z., Ye, W., Terven, J., Bennett, Z., Zheng, Y., Jiang, T., Huang, T.: {MUVA:} {A} new large-scale benchmark for multi-view amodal instance segmentation in the shopping scenario. In: ICCV (2023)

\bibitem{Liao2022PAMI}
Liao, Y., Xie, J., Geiger, A.: {KITTI}-360: {A} novel dataset and benchmarks for urban scene understanding in {2D} and {3D}. IEEE Transactions on Pattern Analysis and Machine Intelligence  (2023)

\bibitem{ling2023panoswin}
Ling, Z., Xing, Z., Zhou, X., Cao, M., Zhou, G.: {PanoSwin:} {A} pano-style swin transformer for panorama understanding. In: CVPR (2023)

\bibitem{liu2024blade}
Liu, Z., Li, Z., Jiang, T.: {BLADE:} {Box-level} supervised amodal segmentation through directed expansion. In: AAAI (2024)

\bibitem{loshchilov2017decoupled}
Loshchilov, I., Hutter, F.: Decoupled weight decay regularization. In: ICLR (2019)

\bibitem{luo2019taking_clan}
Luo, Y., Zheng, L., Guan, T., Yu, J., Yang, Y.: Taking a closer look at domain shift: Category-level adversaries for semantics consistent domain adaptation. In: CVPR (2019)

\bibitem{ma2021densepass}
Ma, C., Zhang, J., Yang, K., Roitberg, A., Stiefelhagen, R.: {DensePASS:} {Dense} panoramic semantic segmentation via unsupervised domain adaptation with attention-augmented context exchange. In: ITSC (2021)

\bibitem{mei2022waymo}
Mei, J., Zhu, A.Z., Yan, X., Yan, H., Qiao, S., Chen, L.C., Kretzschmar, H.: Waymo open dataset: Panoramic video panoptic segmentation. In: ECCV (2022)

\bibitem{mohan2022amodal_panoptic_segmentation}
Mohan, R., Valada, A.: Amodal panoptic segmentation. In: CVPR (2022)

\bibitem{mohan2022perceiving}
Mohan, R., Valada, A.: Perceiving the invisible: Proposal-free amodal panoptic segmentation. IEEE Robotics and Automation Letters  (2022)

\bibitem{nanay2018importance}
Nanay, B.: The importance of amodal completion in everyday perception. i-Perception  (2018)

\bibitem{orhan2022semantic_outdoor}
Orhan, S., Bastanlar, Y.: Semantic segmentation of outdoor panoramic images. Signal, Image and Video Processing  (2022)

\bibitem{poudel2019fastscnn}
Poudel, R.P.K., Liwicki, S., Cipolla, R.: {Fast-SCNN:} {Fast} semantic segmentation network. In: BMVC (2019)

\bibitem{psomas2023simpool}
Psomas, B., Kakogeorgiou, I., Karantzalos, K., Avrithis, Y.: Keep it {SimPool}: {Who} said supervised transformers suffer from attention deficit? In: ICCV (2023)

\bibitem{qi2019amodal_kins}
Qi, L., Jiang, L., Liu, S., Shen, X., Jia, J.: Amodal instance segmentation with {KINS} dataset. In: CVPR (2019)

\bibitem{saha2023edaps}
Saha, S., Hoyer, L., Obukhov, A., Dai, D., Van~Gool, L.: {EDAPS:} {Enhanced} domain-adaptive panoptic segmentation. In: ICCV (2023)

\bibitem{sekkat2020omniscape}
Sekkat, A.R., Dupuis, Y., Vasseur, P., Honeine, P.: The omniscape dataset. In: ICRA (2020)

\bibitem{sekkat2023amodalsynthdrive}
Sekkat, A.R., Mohan, R., Sawade, O., Matthes, E., Valada, A.: {AmodalSynthDrive:} {A} synthetic amodal perception dataset for autonomous driving. arXiv preprint arXiv:2309.06547  (2023)

\bibitem{shen2022panoformer}
Shen, Z., Lin, C., Liao, K., Nie, L., Zheng, Z., Zhao, Y.: {PanoFormer:} {Panorama} transformer for indoor 360{\textdegree} depth estimation. In: ECCV (2022)

\bibitem{shi2023fishdreamer}
Shi, H., Li, Y., Yang, K., Zhang, J., Peng, K., Roitberg, A., Ye, Y., Ni, H., Wang, K., Stiefelhagen, R.: {FishDreamer:} {Towards} fisheye semantic completion via unified image outpainting and segmentation. In: CVPRW (2023)

\bibitem{shi2022unsupervised}
Shi, Y., Ying, X., Zha, H.: Unsupervised domain adaptation for semantic segmentation of urban street scenes reflected by convex mirrors. IEEE Transactions on Intelligent Transportation Systems  (2022)

\bibitem{sun2022amodal_bayesian}
Sun, Y., Kortylewski, A., Yuille, A.: Amodal segmentation through out-of-task and out-of-distribution generalization with a {Bayesian} model. In: CVPR (2022)

\bibitem{tateno2018distortion}
Tateno, K., Navab, N., Tombari, F.: Distortion-aware convolutional filters for dense prediction in panoramic images. In: ECCV (2018)

\bibitem{teng2023360bev}
Teng, Z., Zhang, J., Yang, K., Peng, K., Shi, H., Rei{\ss}, S., Cao, K., Stiefelhagen, R.: {360BEV:} {Panoramic} semantic mapping for indoor bird's-eye view. In: WACV (2024)

\bibitem{tran2022aisformer}
Tran, M., Vo, K., Yamazaki, K., Fernandes, A., Kidd, M., Le, N.: {AISFormer:} {Amodal} instance segmentation with transformer. In: BMVC (2022)

\bibitem{tranheden2021dacs}
Tranheden, W., Olsson, V., Pinto, J., Svensson, L.: {DACS:} {Domain} adaptation via cross-domain mixed sampling. In: WACV (2021)

\bibitem{wang2021hrnet}
Wang, J., Sun, K., Cheng, T., Jiang, B., Deng, C., Zhao, Y., Liu, D., Mu, Y., Tan, M., Wang, X., Liu, W., Xiao, B.: Deep high-resolution representation learning for visual recognition. IEEE Transactions on Pattern Analysis and Machine Intelligence  (2021)

\bibitem{wang2021pvt}
Wang, W., Xie, E., Li, X., Fan, D.P., Song, K., Liang, D., Lu, T., Luo, P., Shao, L.: Pyramid vision transformer: A versatile backbone for dense prediction without convolutions. In: ICCV (2021)

\bibitem{wang2020differential}
Wang, Z., Yu, M., Wei, Y., Feris, R., Xiong, J., Hwu, W., Huang, T.S., Shi, H.: Differential treatment for stuff and things: A simple unsupervised domain adaptation method for semantic segmentation. In: CVPR (2020)

\bibitem{xiao2021amodal_segmentation_prior}
Xiao, Y., Xu, Y., Zhong, Z., Luo, W., Li, J., Gao, S.: Amodal segmentation based on visible region segmentation and shape prior. In: AAAI (2021)

\bibitem{xie2021segformer}
Xie, E., Wang, W., Yu, Z., Anandkumar, A., {\'{A}}lvarez, J.M., Luo, P.: {SegFormer:} {Simple} and efficient design for semantic segmentation with transformers. In: NeurIPS (2021)

\bibitem{xu2019semantic}
Xu, Y., Wang, K., Yang, K., Sun, D., Fu, J.: Semantic segmentation of panoramic images using a synthetic dataset. In: SPIE (2019)

\bibitem{yang2019can}
Yang, K., Hu, X., Bergasa, L.M., Romera, E., Huang, X., Sun, D., Wang, K.: Can we {PASS} beyond the field of view? {Panoramic} annular semantic segmentation for real-world surrounding perception. In: IV (2019)

\bibitem{yang2019pass}
Yang, K., Hu, X., Bergasa, L.M., Romera, E., Wang, K.: {PASS:} {Panoramic} annular semantic segmentation. IEEE Transactions on Intelligent Transportation Systems  (2020)

\bibitem{yang2020ds}
Yang, K., Hu, X., Chen, H., Xiang, K., Wang, K., Stiefelhagen, R.: {DS-PASS:} {Detail-sensitive} panoramic annular semantic segmentation through {SwaftNet} for surrounding sensing. In: IV (2020)

\bibitem{yang2020omnisupervised}
Yang, K., Hu, X., Fang, Y., Wang, K., Stiefelhagen, R.: Omnisupervised omnidirectional semantic segmentation. IEEE Transactions on Intelligent Transportation Systems  (2022)

\bibitem{yang2021context}
Yang, K., Hu, X., Stiefelhagen, R.: Is context-aware {CNN} ready for the surroundings? {Panoramic} semantic segmentation in the wild. IEEE Transactions on Image Processing  (2021)

\bibitem{yang2021capturing}
Yang, K., Zhang, J., Rei{\ss}, S., Hu, X., Stiefelhagen, R.: Capturing omni-range context for omnidirectional segmentation. In: CVPR (2021)

\bibitem{ye2020universal}
Ye, Y., Yang, K., Xiang, K., Wang, J., Wang, K.: Universal semantic segmentation for fisheye urban driving images. In: SMC (2020)

\bibitem{yogamani2019woodscape}
Yogamani, S.K., Witt, C., Rashed, H., Nayak, S., Mansoor, S., Varley, P., Perrotton, X., O'Dea, D., P{\'{e}}rez, P., Hughes, C., Horgan, J., Sistu, G., Chennupati, S., Uric{\'{a}}r, M., Milz, S., Simon, M., Amende, K.: {WoodScape:} {A} multi-task, multi-camera fisheye dataset for autonomous driving. In: ICCV (2019)

\bibitem{yu2023osrt}
Yu, F., Wang, X., Cao, M., Li, G., Shan, Y., Dong, C.: {OSRT:} {Omnidirectional} image super-resolution with distortion-aware transformer. In: CVPR (2023)

\bibitem{yu2023panelnet}
Yu, H., He, L., Jian, B., Feng, W., Liu, S.: {PanelNet:} {Understanding} 360 indoor environment via panel representation. In: CVPR (2023)

\bibitem{yu2022metaformer}
Yu, W., Luo, M., Zhou, P., Si, C., Zhou, Y., Wang, X., Feng, J., Yan, S.: {MetaFormer} is actually what you need for vision. In: CVPR (2022)

\bibitem{yuan2021robust_instance_segmentation}
Yuan, X., Kortylewski, A., Sun, Y., Yuille, A.: Robust instance segmentation through reasoning about multi-object occlusion. In: CVPR (2021)

\bibitem{yue2021prototypical}
Yue, X., Zheng, Z., Zhang, S., Gao, Y., Darrell, T., Keutzer, K., Vincentelli, A.S.: Prototypical cross-domain self-supervised learning for few-shot unsupervised domain adaptation. In: CVPR (2021)

\bibitem{zhang2021deeppanocontext}
Zhang, C., Cui, Z., Chen, C., Liu, S., Zeng, B., Bao, H., Zhang, Y.: {DeepPanoContext:} {Panoramic} {3D} scene understanding with holistic scene context graph and relation-based optimization. In: ICCV (2021)

\bibitem{zhang2021transfer}
Zhang, J., Ma, C., Yang, K., Roitberg, A., Peng, K., Stiefelhagen, R.: Transfer beyond the field of view: Dense panoramic semantic segmentation via unsupervised domain adaptation. IEEE Transactions on Intelligent Transportation Systems  (2022)

\bibitem{zhang2022bending}
Zhang, J., Yang, K., Ma, C., Rei{\ss}, S., Peng, K., Stiefelhagen, R.: Bending reality: Distortion-aware transformers for adapting to panoramic semantic segmentation. In: CVPR (2022)

\bibitem{zhang2022behind}
Zhang, J., Yang, K., Shi, H., Rei{\ss}, S., Peng, K., Ma, C., Fu, H., Torr, P.H.S., Wang, K., Stiefelhagen, R.: Behind every domain there is a shift: Adapting distortion-aware vision transformers for panoramic semantic segmentation. IEEE Transactions on Pattern Analysis and Machine Intelligence  (2024)

\bibitem{zhang2022UniDAPS}
Zhang, J., Huang, J., Lu, S.: Hierarchical mask calibration for unified domain adaptive panoptic segmentation. In: CVPR (2023)

\bibitem{zhang2019learning_distance}
Zhang, Z., Chen, A., Xie, L., Yu, J., Gao, S.: Learning semantics-aware distance map with semantics layering network for amodal instance segmentation. In: MM (2019)

\bibitem{zheng2023look_neighbor}
Zheng, X., Pan, T., Luo, Y., Wang, L.: Look at the neighbor: Distortion-aware unsupervised domain adaptation for panoramic semantic segmentation. In: ICCV (2023)

\bibitem{zheng2023both_style_distortion}
Zheng, X., Zhu, J., Liu, Y., Cao, Z., Fu, C., Wang, L.: Both style and distortion matter: Dual-path unsupervised domain adaptation for panoramic semantic segmentation. In: CVPR (2023)

\bibitem{zheng2023complementary_bidirectional}
Zheng, Z., Lin, C., Nie, L., Liao, K., Shen, Z., Zhao, Y.: Complementary bi-directional feature compression for indoor 360{\textdegree} semantic segmentation with self-distillation. In: WACV (2023)

\bibitem{zhou2022fan}
Zhou, D., Yu, Z., Xie, E., Xiao, C., Anandkumar, A., Feng, J., {\'{A}}lvarez, J.M.: Understanding the robustness in vision transformers. In: ICML (2022)

\bibitem{zhu2017semantic_amodal_segmentation}
Zhu, Y., Tian, Y., Metaxas, D., Doll{\'a}r, P.: Semantic amodal segmentation. In: CVPR (2017)

\bibitem{zou2019crst}
Zou, Y., Yu, Z., Liu, X., Kumar, B.V.K.V., Wang, J.: Confidence regularized self-training. In: ICCV (2019)

\end{thebibliography}

\end{document}